\documentclass[journal]{IEEEtran}
\usepackage{amsmath,amsfonts}
\usepackage{algorithmic}
\usepackage[colorlinks=true,
            linkcolor=blue,
            citecolor=blue,
            urlcolor=blue]{hyperref}

\usepackage{algorithm}
\usepackage{array}
\usepackage[caption=false,font=normalsize,labelfont=sf,textfont=sf]{subfig}
\usepackage{textcomp}
\usepackage{stfloats}
\usepackage{url}
\usepackage{verbatim}
\usepackage{graphicx}
\usepackage{cite}
\usepackage{multirow}
\usepackage{soul}
\usepackage[T1]{fontenc}
\input{glyphtounicode}
\newcommand{\fig}[1]{Fig.~\ref{fig:#1}} 
\newcommand{\Table}[1]{Table~\ref{tab:#1}}
\newcommand{\secref}[1]{Sec.~\ref{sec:#1}}
\usepackage{xcolor}
\usepackage{booktabs}
\newcounter{todocounter}

\newcommand{\revise}[1]{{#1}}

\setstcolor{cyan}
\setul{}{3pt} 
\newcommand{\gst}[1]{{}}

\begin{document}
\bstctlcite{BSTcontrol} 

\title{\gst{AME-2:} Agile and Generalized Legged Locomotion via\\ Attention-Based Neural Map Encoding}

\author{Chong Zhang$^{123*}$, Victor Klemm$^{1}$, Fan Yang$^{1}$, Marco Hutter$^{1}$\\
Website: \url{https://sites.google.com/leggedrobotics.com/ame-2}
\thanks{$^{1}$Robotic Systems Lab, ETH Zurich, Switzerland}
\thanks{$^{2}$Secure, Reliable, and Intelligent Systems Lab, ETH Zurich, Switzerland}%
\thanks{$^{3}$ETH AI Center, Switzerland}%
\thanks{$^{*}$Corresponding: chong.zhang@ai.ethz.ch}}

\markboth{Journal Submission}%
{Anonymous \MakeLowercase{\textit{et al.}}: Anonymous Article Using IEEEtran.cls for IEEE Journals}


\maketitle

\begin{abstract} 
Achieving agile and generalized legged locomotion across terrains requires tight integration of perception and control, especially under occlusions and sparse footholds. 
Existing methods have demonstrated agility on parkour courses but often rely on end-to-end sensorimotor models with limited generalization and interpretability. By contrast, methods targeting generalized locomotion typically exhibit limited agility and struggle with visual occlusions. 
We introduce \gst{AME-2,} a unified reinforcement learning (RL) framework for agile and generalized locomotion that incorporates a novel attention-based map encoder in the control policy. 
This encoder extracts local and global mapping features and uses attention mechanisms to focus on salient regions, producing an interpretable and generalized embedding for RL-based control.
We further propose a learning-based mapping pipeline that provides fast, uncertainty-aware terrain representations robust to noise and occlusions, serving as policy inputs. It uses neural networks to convert depth observations into local elevations with uncertainties, and fuses them with odometry. The pipeline also integrates with parallel simulation so that we can train controllers with online mapping, aiding sim-to-real transfer.
We validate \revise{our framework} with the proposed mapping pipeline on a quadruped and a biped robot, and the resulting controllers demonstrate strong agility and generalization to unseen terrains in simulation and in real-world experiments.
\end{abstract} 


\section{Introduction}
\IEEEPARstart{A}{gile} and generalized locomotion is essential for legged robots to operate reliably in diverse real-world environments. It requires tightly coupled real-time perception and control, robustness to sensor noise and occlusions, agility through whole-body control, and precise behavior on terrains with sparse footholds. Designing such perceptive locomotion systems that are simultaneously agile and able to generalize across diverse terrains remains a critical challenge.

Classical pipelines address perceptive locomotion by combining model-based control with explicit mapping and state estimation~\cite{fankhauser2018robust, jenelten2022tamols, fahmi2022vital, grandia2023perceptive, akizhanov2024learning, 9035046, mastalli2022agile, 9134750}. Typically, they maintain a state estimator~\cite{8772165}, build elevation maps from sensor data~\cite{8392399, takamapping}, and use model-based planning and control methods to command the robot based on this mapping information. These approaches have proven effective for deliberate walking and carefully planned maneuvers on moderately structured terrain, where accurate state estimation and dense, slowly changing maps can be maintained. However, their reliance on precise models and often deterministic optimal control makes them sensitive to errors in estimation and mapping, especially under visual occlusions and violations of modeling assumptions~\cite{grandia2023perceptive}. Elevation maps are often updated at lower rates than the control loop and heuristically filtered~\cite{takamapping, jenelten2022tamols, grandia2023perceptive}, which often requires per-terrain tuning to ensure safety and makes it difficult to obtain a general solution for diverse terrains that demand agile motions. Moreover, the optimization and planning components in these pipelines tend to be computationally heavy, which can limit agility on real robots. These limitations motivate learning-based approaches such as RL that can better exploit raw sensor data, operate under uncertainty, and produce more dynamic behaviors.

Learning-based methods, in particular RL, have recently shown strong potential for legged locomotion~\cite{ha2025learning}. One line of work combines RL with classical state estimation and mapping pipelines, using elevation maps or related representations as policy inputs~\cite{rudin2022learning, miki2022learning, he2025attention, videomimic, long2025learning, ren2025vb, wang2025beamdojo, nikitagoalreaching, jenelten2024dtc, gangapurwala2022rloc, dong2025marg, chen2025learning, 10610271, zhang2024risky}. These methods can improve robustness and generalization compared to purely model-based controllers, but they also inherit the computational cost and failure modes of the underlying estimation and mapping systems~\cite{adv, lee2024learning}. In simulation, noise is typically injected into both the state estimates and the mapping observations. A common strategy is to assume a fully observed egocentric map during training and then tune the real-world mapping stack to approximate this assumption~\cite{rudin2022learning, jenelten2024dtc, he2025attention}, which simplifies training but can make performance sensitive when occlusions occur or when agile whole-body contacts violate estimator assumptions. Another strategy is to heavily randomize state-estimation and mapping noise so that the controller becomes robust and conservative under mapping uncertainties in dense terrains~\cite{miki2022learning}, yet such policies tend to struggle on terrains that require very high agility or precise foothold placement.

More recently, neural network–based mapping has been explored, where networks are trained in simulation to reconstruct maps that can be used as policy inputs~\cite{sun2025dpl, yang2025agile, hoeller2022neural, hoeller2024anymal, yu2024walking, duan2024learning}. While this yields efficient mapping suitable for real-time control, the learned mappings are usually fitted to particular terrain distributions and sensor configurations, offer limited generalization to unseen terrains, and do not explicitly model uncertainties such as occlusions.

Another line of work directly maps raw exteroceptive perception and proprioception to actions~\cite{pmlr-v205-agarwal23a, cheng2024extreme, yanglearning, zhuang2025humanoid, 11127928, ben2025gallant, 10678805, chanesane2025soloparkour, kareer2023vinl, 11128762, li2025kivi,11203201, wang2025more}. These sensorimotor policies have demonstrated highly agile behaviors on challenging courses~\cite{cheng2024extreme, zhuang2025humanoid,chanesane2025soloparkour}, but typically exhibit limited generalization beyond the training environments and offer little interpretability, since terrain reasoning is implicitly encoded in the policy. More recent sensorimotor generalist models~\cite{rudin2025parkour} distill a single policy from multiple expert controllers to achieve strong performance, yet such methods still show limited generalization to unseen terrains and often require finetuning on new terrains, as evaluated in our benchmarks (\secref{benchmark}). Overall, existing RL-based approaches tend to trade off agility, generalization, mapping efficiency, and interpretability rather than addressing these requirements within a unified framework.

\begin{figure*}[ht]
      \centering
    \includegraphics[width=0.96\linewidth]{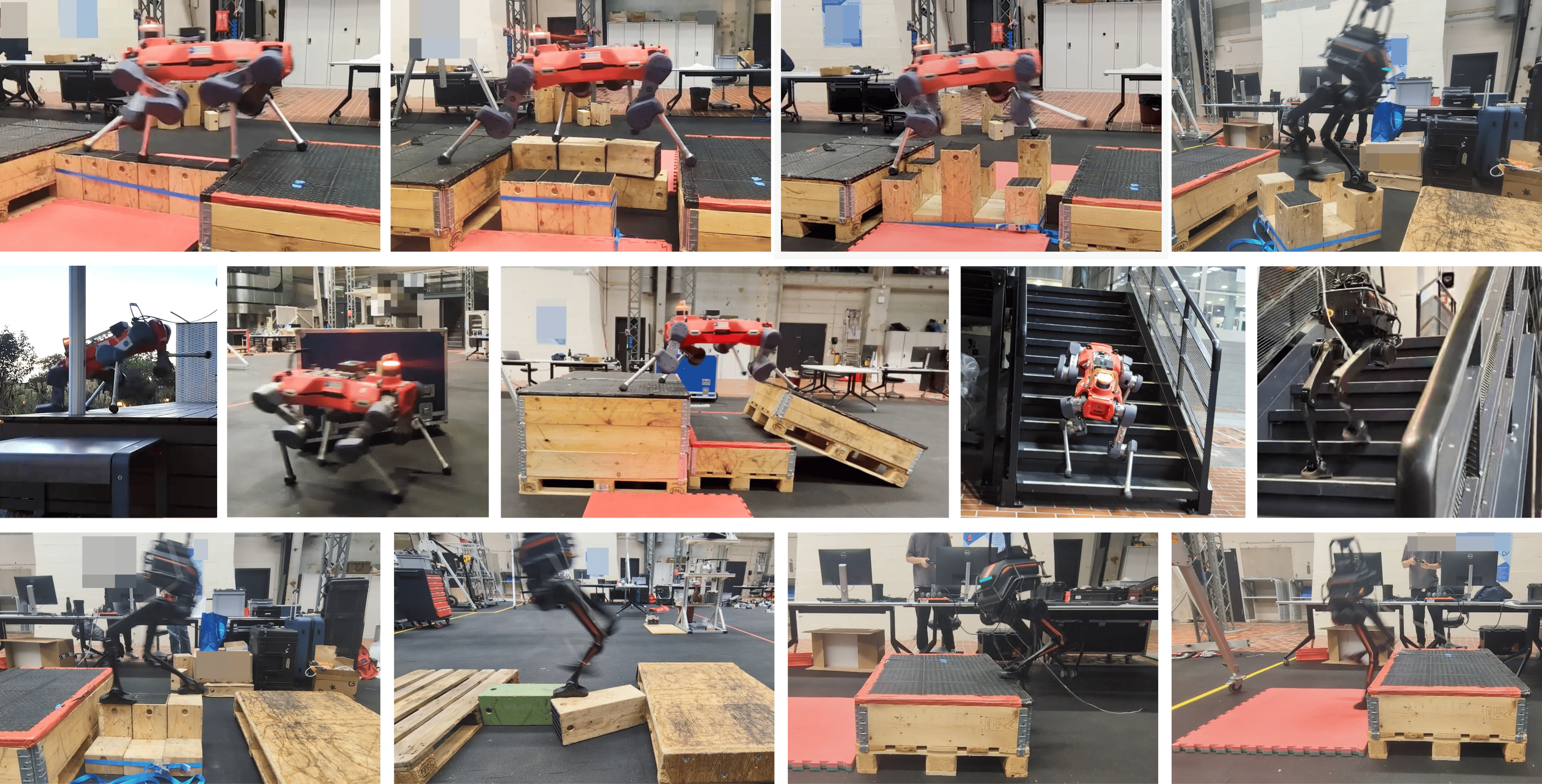} 
    \vspace{-0.35cm}
    \caption{Our method enables agile and generalized legged locomotion across diverse terrains with onboard sensing and computation.}
    \vspace{-0.5cm}
    \label{fig:fig1}  
\end{figure*}

In this work, we present \gst{AME-2,} a unified RL framework for perceptive legged locomotion with an \textbf{A}ttention-based \textbf{M}ap \textbf{E}ncoding architecture trained jointly with the controller. \revise{A proposed} AME-2 encoder, built upon the design in~\cite{he2025attention}, first extracts local features (pixel-wise representation of terrain details) and global features (capturing global terrain context) from the elevation map, then uses the global features together with proprioception to assign attention weights to the local features. The resulting weighted local features are concatenated with the global features and proprioception to form a terrain-aware representation for policy learning. This design allows the policy to downweigh local regions that are less relevant for the current task, improving generalization to new terrains. Because the attention is conditioned on global terrain context, the policy can also learn distinct attention and motion patterns across different terrains. To enable learning agile controllers with this representation, we adopt a goal-reaching locomotion formulation from prior works~\cite{nikitagoalreaching, zhang2024risky, he2024agile, ben2025gallant, hoeller2024anymal, 10611254} and adapt it so that the same reward functions and training settings can be used for different robots. Together, the AME-2 encoder and unified training formulation enable training a single generalist policy with terrain-aware skills, which in our experiments exhibits strong generalization to unseen terrains while maintaining high agility.

To enable real-world deployment of agile and generalized locomotion controllers, we develop a learning-based elevation mapping pipeline to remove the dependency on classical mapping stacks while maintaining generalization. The pipeline projects depth images into local grids and uses a lightweight neural network trained via Bayesian learning~\cite{kendall2017uncertainties} to predict local elevations with per-cell uncertainty estimates. These local maps are fused into a consistent global frame with odometry, providing a fast and uncertainty-aware representation that accounts for occlusions and sensor noise. We then query egocentric elevations and associated uncertainties as the map inputs for the controller. This design is inspired by~\cite{jung2025uncertainty}, which fuses per-frame map predictions instead of training temporal networks for off-road navigation, thereby reducing data requirements and mitigating overfitting. Furthermore, we synthesize random terrain data, including many terrains that are not feasible for the robots, to broaden the training distribution for the local map predictor and improve generalization.

The same mapping pipeline runs both in parallel simulation and on the real robots, enabling online mapping during training rather than relying on idealized or hand-tuned maps. For training efficiency, we adopt a teacher–student scheme~\cite{rudin2025parkour, wang2025integrating}: a teacher policy is first trained with ground-truth elevation maps, and a student policy with the same AME-2 architecture is then trained using the proposed mapping under teacher supervision alongside RL. The resulting student policy is directly deployable, operates with the lightweight, uncertainty-aware mapping in the loop, and retains the generalization and agility of the teacher.

We evaluate \revise{our method} on an ANYmal-D quadruped~\cite{hutter2016anymal} and a LimX TRON1 biped~\cite{limx_tron1_2025} both in simulation and on real hardware. As shown in \fig{fig1}, the resulting controllers exhibit strong agility and robust generalization to a wide range of terrains. The proposed mapping pipeline produces high-quality elevation maps in real time, supporting these agile motions with onboard perception and computation.

The main contributions of this work are summarized as follows:
\begin{enumerate}
    \item We propose \gst{\textbf{AME-2},} \textbf{a unified RL framework} with an attention-based map encoder to achieve agile and generalized legged locomotion.
    \item We develop a \textbf{lightweight, uncertainty-aware elevation mapping} pipeline that generalizes to diverse terrains, explicitly models occlusions and noise, and helps bridge the sim-to-real gap.
    \item We use a \textbf{teacher-student scheme} that yields deployable controllers using the learned mapping while retaining agility, generalization, and interpretability.
    \item We demonstrate a \textbf{state-of-the-art combination of agility and generalization} with a quadruped and a biped robot across diverse challenging terrains under the same training setup.
\end{enumerate}

\section{Related Works}

\subsection{Perceptive Locomotion with Explicit Mapping}

Early explorations of perceptive locomotion use model-based methods with either prebuilt maps or classical online mapping. Most of them first perform foothold or trajectory planning and then apply a model-based tracking controller~\cite{fankhauser2018robust, jenelten2022tamols, fahmi2022vital, akizhanov2024learning, 9035046, mastalli2022agile, 9134750}, while some directly optimize a reactive controller over feasible footholds~\cite{grandia2023perceptive}. Although these methods offer strong guarantees and can generalize well under their modeling assumptions, their real-world performance is limited by model mismatch, state-estimation uncertainties, mapping errors, and computational burden.

To address these limitations, RL emerges as an alternative for perceptive locomotion~\cite{ha2025learning}. One line of work keeps a strong model-based component and uses RL mainly as an add-on~\cite{jenelten2024dtc,gangapurwala2022rloc,lee2024integrating}. DTC~\cite{jenelten2024dtc} augments a model-based planner with an RL tracking controller to improve robustness while maintaining generalization, but the planner remains sensitive to mapping uncertainties and constrains agility through its modeling assumptions and computational cost. RLOC~\cite{gangapurwala2022rloc} instead uses a model-based controller to track an RL planner, yet the overall performance is capped by the controller.

Other approaches remove the model-based modules entirely while still relying on explicit maps. When given prebuilt maps, RL controllers can exhibit highly agile motions~\cite{zhang2024risky}. With classical online mapping, they outperform model-based counterparts in terms of agility and robustness~\cite{rudin2022learning, miki2022learning, he2025attention, videomimic, long2025learning, ren2025vb, wang2025beamdojo,  dong2025marg, chen2025learning}, but their agility remains limited by the update rate and quality of the mapping. In addition, many such controllers suffer from generalization issues, as they tend to overfit to the training terrains~\cite{jenelten2024dtc}. A recent work~\cite{he2025attention} shows that RL can achieve generalized locomotion across diverse terrains, but its agility is still constrained by the online classical mapping stack and the learning framework.

More recently, RL-based methods have begun to use learned maps, where neural networks reconstruct egocentric terrains for the controller in real time~\cite{sun2025dpl, yang2025agile, hoeller2022neural, hoeller2024anymal, yu2024walking, duan2024learning}. These approaches demonstrate strong agility and successful real-world deployment, but the learned maps often fit the training distribution and show limited generalization in unseen, unstructured environments. In contrast, our work combines a fully RL-based controller with a learned mapping module designed for generalization. As a result, our policies achieve agility comparable to previous state of the art while enabling generalized locomotion across diverse terrains.

\subsection{Perceptive Locomotion with Raw Sensor Data}

Perceptive locomotion with raw sensor data seeks to bypass the limitations of explicit mapping. Instead of constructing and maintaining a map, some works directly train policies from camera images~\cite{pmlr-v205-agarwal23a, cheng2024extreme, yanglearning, zhuang2025humanoid, 11127928, 10678805, 11128762,11203201, chanesane2025soloparkour, kareer2023vinl,  li2025kivi, wang2025more} or lidar point clouds~\cite{ben2025gallant}. Early approaches in this direction still separate perception and control: a visual module outputs footholds or trajectories, which are then tracked by model-based controllers~\cite{yu2021visuallocomotion}. Between explicit mapping and fully end-to-end policies, other works learn intermediate representations from raw sensors (such as ray distances~\cite{he2024agile} and neural volumetric memory~\cite{yang2023neural}) that remain cheap to compute while enabling fast control. 

Many recent methods adopt a monolithic design, where a single neural network maps raw sensor streams and proprioception directly to joint commands. Within this family, some distill deployable student policies from privileged teacher policies — often using maps as privileged inputs — to achieve efficient and stable learning~\cite{pmlr-v205-agarwal23a, cheng2024extreme, zhuang2025humanoid, 11203201, chanesane2025soloparkour, kareer2023vinl, rudin2025parkour}, while others learn from scratch with the help of representation learning techniques such as world models~\cite{11128762} or privileged-information reconstruction~\cite{10678805, li2025kivi}. 
The current state of the art in agile generalist locomotion~\cite{rudin2025parkour} trains multiple mapping-based teacher controllers and distills them into a single generalist student policy that operates directly on raw depth images. This student policy combines the skills of the teachers and can be finetuned efficiently on new terrains. However, its zero-shot generalization to previously unseen environments is limited.

In contrast, our work also trains neural networks from raw sensor data, but uses them exclusively within the mapping process rather than in a monolithic perception-to-action policy. Because our learned mapping is lightweight and well suited to massive parallelization in simulation, we can run the full mapping loop during training and then deploy exactly the same mapping at test time, preserving both speed and consistency. Without training a monolithic policy that directly maps raw sensor data to joint actions, we still obtain an agile controller while also achieving strong generalization across diverse terrains.

\subsection{Learning-Based Mapping}

Learning-based mapping aims to retain the structure and interpretability of explicit maps while overcoming the computational and modeling limitations of classical mapping. Classical mapping pipelines combine geometric projection, hand-tuned filtering, and probabilistic fusion to produce high-quality maps, and can handle a wide range of complex terrains~\cite{8392399, takamapping, dong2025marg, chen2025learning}. However, these methods are often too slow for agile locomotion and require heuristic, terrain-specific filtering to cope with occlusions and sensor noise. 

To enable faster control, several works directly train terrain reconstruction modules in or after the policy training loop~\cite{sun2025dpl, yang2025agile, hoeller2022neural, hoeller2024anymal, yu2024walking, duan2024learning}. These learned mappings can infer local geometry quickly enough to support agile controllers, but are usually trained on a narrow distribution of environments, showing limited generalization beyond the training terrains.

Recent off-road navigation approaches take a different route to obtain uncertainty-aware neural mapping with improved accuracy and data efficiency. In particular, \cite{jung2025uncertainty} leverages neural processes~\cite{kim2019attentive} to model terrain elevation: a network predicts per-frame local elevation maps together with associated uncertainties, which are then fused over time using odometry. By predicting local maps with uncertainty instead of learning a fully end-to-end sensor-to-map model, this approach reduces the data requirements for accurate estimation and improves generalization in complex off-road environments. However, the computational costs of the models in \cite{jung2025uncertainty} are too high (each environment takes $>13$ GB GPU memory) and cannot be deployed with thousands of parallel simulation environments.

Inspired by these ideas, we design a learning-based elevation mapping module tailored for agile legged locomotion. We project depth-camera point clouds into local elevation grids and train lightweight, robust networks to predict local elevation and its uncertainty from these inputs, which are then fused over time using odometry. This design suppresses sensor noise and naturally models occlusions as regions of high uncertainty. We further improve generalization through extensive randomization in a custom data pipeline. The resulting module is lightweight enough to run in massively parallel simulation and fast enough to support highly agile motions in deployment, while providing an uncertainty-aware terrain representation for our controllers.

\section{System Overview}

\begin{figure*}[ht]
    \centering
    \includegraphics[width=0.9\linewidth]{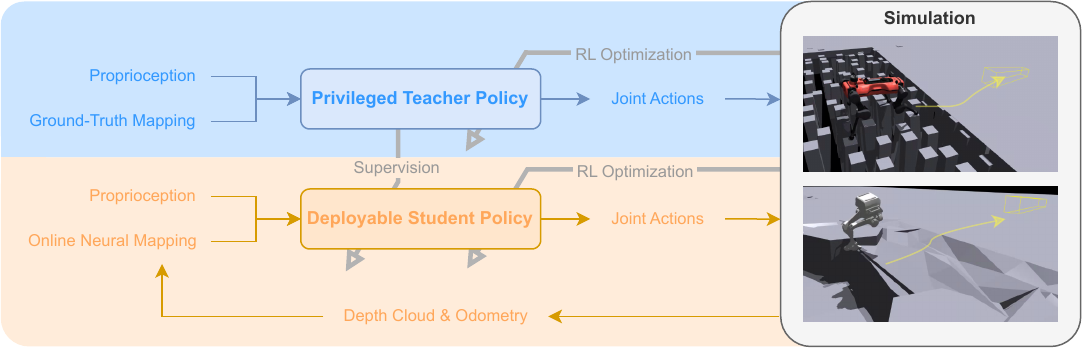}
    \vspace{-0.2cm}
    \caption{An overview of our system. We use RL to train a teacher policy with ground-truth mapping in simulation, and a student policy under the teacher's supervision with our proposed neural mapping. The policies output joint-level actions that actuate the robots to reach position and heading goals, as illustrated by the yellow wireframes.}
    \vspace{-0.5cm}
    \label{fig:system}
\end{figure*}

\subsection{Formulation}

As illustrated in \fig{system}, we use RL to train terrain-aware locomotion policies that reach position and heading goals. We formulate the problem as a partially observable Markov decision process (POMDP) and optimize the policies using PPO~\cite{schulman2017proximal} in parallel simulation~\cite{makoviychuk2021isaac}. By maximizing goal-reaching rewards (detailed in \secref{control}), the policies learn agile locomotion skills that enable them to traverse diverse terrains. We first train a privileged teacher policy {via asymmetric actor–critic} and then transfer the skills to a deployable student policy, which is also detailed in \secref{control}.

\subsection{Controller Inputs and Outputs}
Our controllers run at 50 Hz. The policy actions $a$ are joint PD targets~\cite{xbpeng_choiceofA} tracked at 400 Hz on real hardware. 

Proprioception observations include base linear velocity $v_b$ (only accessible to the teacher), base angular velocity $\omega_b$, projected gravity $g_b$, joint positions $q$, joint velocities $\dot{q}$, previous actions, and goal commands $c$.

Ground-truth mapping represents each point in the egocentric elevation grid by its 3D coordinates $(x, y, z)$. Our neural mapping augments this with an uncertainty channel and produces a 4D representation $(x, y, z, u)$ for each point, where $u$ is the uncertainty metric.

\section{Locomotion Control}
\label{sec:control}
\subsection{Policy Architecture and AME-2 Encoder}

\begin{figure*}
    \centering
    \includegraphics[width=0.92\linewidth]{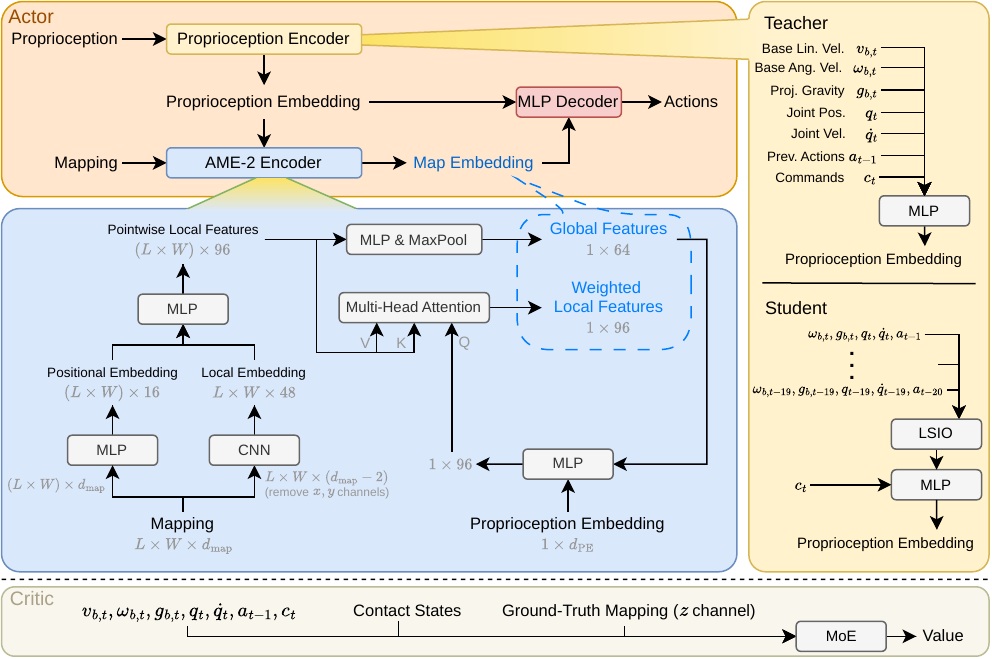}
    \vspace{-0.2cm}
    \caption{An illustration of our \revise{network} architecture. \textbf{Left top:} The general abstract of our policies. Proprioceptive observations are encoded into a proprioception embedding, which is used together with the mapping as the inputs of the AME-2 encoder to produce the map embedding. The proprioception embedding and the map embedding are then concatenated and fed into an MLP to generate actions. \textbf{Left {middle}:} The AME-2 encoder design. We extract both local features and global features from the map, and then use the global features and the proprioception embedding to produce the attention-weighted local features. $L$ and $W$ are the length and width of the map, $d_{map}$ is the dimension of map representation (3 for the teacher, 4 for the student), $d_{PE}$ is the dimension of proprioception embedding. \textbf{Right:} The proprioception encoder designs. We have different designs for teacher and student policies to facilitate sim-to-real transfer, while both designs can fit into our overall architecture. {\textbf{Bottom}: The critic is a mixture-of-experts model, which is powerful for function fitting yet efficient to train.}}
    \vspace{-0.6cm}
    \label{fig:arch}
\end{figure*}

The general policy architecture is illustrated in \fig{arch}, with the AME-2 encoder serving as the core feature extractor for the mapping input. We have a proprioception encoder to embed proprioceptive observations, and the AME-2 encoder to embed the map observations based on the proprioception embedding. We feed both the proprioception embedding and the map embedding through a multilayer perceptron (MLP) to output the actions.

The AME-2 encoder first extracts local map features with a convolutional neural network (CNN) and computes a positional embedding for each point with an MLP. These are then fused by another MLP to obtain pointwise local features. Next, an additional MLP followed by max pooling over the pointwise features produces global features that capture the overall terrain context. We combine these global features with a proprioceptive embedding through an MLP to obtain a query vector, which is used in a multi-head attention (MHA) module~\cite{vaswani2017attention} with the pointwise local features serving as keys and values. This yields a weighted local feature embedding that focuses on important terrain regions based on the current proprioceptive state and global context. Finally, the global features and weighted local features are concatenated to form the map embedding fed into the policy’s action decoder.

This design builds on the attention-based map encoder in previous work \cite{he2025attention}, which we refer to as AME-1 for clarity. Unlike the AME-1 encoder, our AME-2 encoder additionally computes global features and uses them to weigh the local features. This enables more generalized locomotion over complex terrains, where motion patterns should vary with the terrain. We demonstrate the resulting performance gap between AME-1 and AME-2 encoders in \secref{results} and \secref{benchmark}.

The choice of proprioception encoder in our architecture is flexible, as long as it produces an effective proprioception embedding. Hence, we use different designs for the teacher and the student. For the teacher, since ground-truth proprioceptive observations are available, we use a plain MLP over them to obtain the proprioception embedding. For the student, because of various environmental uncertainties and following lessons from previous works~\cite{zhang2025wococo, he2025omnih2o, li2024reinforcement}, we stack the proprioceptive observations (except base linear velocities and commands) from the past 20 steps and use Long-Short I/O (LSIO)~\cite{li2024reinforcement} to obtain a temporal embedding of both robot states and environment dynamics. This temporal embedding and the commands are then jointly fed into an MLP to produce the student’s proprioception embedding.

\subsection{Asymmetric Actor Critic}

We adopt asymmetric actor–critic training~\cite{DBLP:conf/rss/PintoAWZA18}. For the critic, {as shown in \fig{arch}, }we do not use the same attention-based design as in the actor, because the critic does not need {to be deployed or }generalize beyond the training terrains, and optimizing an MHA module with $(L \times W)$ local feature inputs is costly. Instead, we use the mixture-of-experts (MoE) design from~\cite{chen2025gmt}, which is powerful for function fitting yet much more computationally efficient to optimize. Note that, although we can train a generalized teacher with our policy architecture and an MoE critic, an MoE actor does not yield a generalized teacher, as shown in \secref{benchmark}.

{
During teacher training, the teacher actor receives noiseless proprioceptive observations and the ground-truth map. The critic receives the same inputs, with the additional privileged contact state of each link, to provide more informative training signals.
We do not provide contact states to the teacher actor because a robust teacher can already be trained without them}, and a larger information gap between teacher and student can hurt student training~\cite{messikommer2025studentinformed}. 
We also apply left-right symmetry augmentation from \cite{hoeller2024anymal} to the critic to improve sample efficiency and motion style, but not to the actor to avoid additional computational cost of actor optimization.
The critic design is shared for the teacher and the student.

\subsection{Teacher-Student RL}

We use Teacher–Student RL~\cite{lee2020learning, miki2022learning, rudin2025parkour, wang2025integrating} to facilitate sim-to-real transfer. Although directly training a student policy with our mapping pipeline is also possible, simulation can run at only about half the speed compared to using ground-truth mapping and requires significantly more GPU memory, making it computationally inefficient for us.

Our student training objective linearly combines the RL losses from PPO, the action distillation losses from~\cite{wang2025integrating}, and a representation loss given by the mean squared error between the teacher and student map embeddings. In practice, we also disable the PPO surrogate loss during the first few iterations while using a large learning rate. These design choices enable stable, efficient, and well-aligned student training, which we will ablate in \secref{benchmark}.

\subsection{Environments} 

\subsubsection{Rewards}

\begin{table*}[ht]
\centering
\caption{Rewards for Controller Training.}
\begin{tabular}{lll>{\raggedleft\arraybackslash}p{6cm}}
\hline
\textbf{Reward Term} & \textbf{Expression} & \textbf{Weight ($\times d\tau^*$)} & \textbf{Notes} \\
\hline
\textit{Task Rewards} &  &  & \\ \cline{1-1}
Position Tracking  & Eq.~(\ref{eq:rew_pos_tracking}) & $100$ & \\
Heading Tracking  &  Eq.~(\ref{eq:rew_yaw_tracking})  & $50$ &  \\
Moving to Goal & Eq.~(\ref{eq:rew_move}) & $5$ & \\
Standing at Goal & Eq.~(\ref{eq:rew_stand}) & $5$ & \\ \hline
\textit{Regularization and Penalties} & & & \\ \cline{1-1}
Early Termination & $1$ if early termination triggered & $-10 / d\tau $ & \\
Undesired Events  & $1$ for each undesired event & $-1$ & \\
Base Roll Rate  &  $[\omega_b]_x^2$ & $-0.1$ & \\
Joint Regularization  & $\lVert \dot{q} \rVert^2 + 0.01\lVert \tau \rVert^2 + 0.001\lVert \ddot{q} \rVert^2$ & $-0.001$ & \footnotesize{$\tau$ denotes the joint torques} \\
Action Smoothness  & $\lVert a_t - a_{t-1} \rVert^2$  & $-0.01$ & \\
Link Contact Forces & $ \lVert \max{(F_{\rm con}- G, \mathbf{0}) \rVert ^ 2}$ & $-0.00001$ &  $F_{\rm con}$ denotes the contact forces for each link and $G$ is the robot weight \\
Link Acceleration & $\sum_l \lVert \dot{v}_l \rVert $ & $-0.001$ & $v_l$ is the velocity of link $l$; summing over all links \\ 
\hline
\textit{Simulation Fidelity} & & & summing over all joints \\ \cline{1-1}
Joint Position Limits & $\sum_{j} \max(0, q_j - 0.95 q^{\max}_j, 0.95 q^{\min}_j - q_j)
$ & $-1000$ & $q_j^{\max}$ and $q_j^{\min}$ are position limits for joint $j$  \\
Joint Velocity Limits & $\sum_{j} \max(0, |\dot{q}_j| - 0.9 \dot{q}^{\max}_j)$ & $-1$ & $ \dot{q}^{\max}_j$ is the velocity limit for joint $j$ \\ 
Joint Torque Limits & $\sum_{j} \max(0, |\tau_j| - 0.8 \tau^{\max}_j)$ & $-1$ & $\tau^{\max}_j$ is the torque limit for joint $j$\\ 

\hline
\multicolumn{3}{l}{\footnotesize $^*d\tau=0.02$ sec is the policy interval.} \\
\end{tabular}
\vspace{-0.5cm}
\label{tab:rewards}
\end{table*}

\begin{figure*}[ht]
    \centering
    \includegraphics[width=0.93\linewidth]{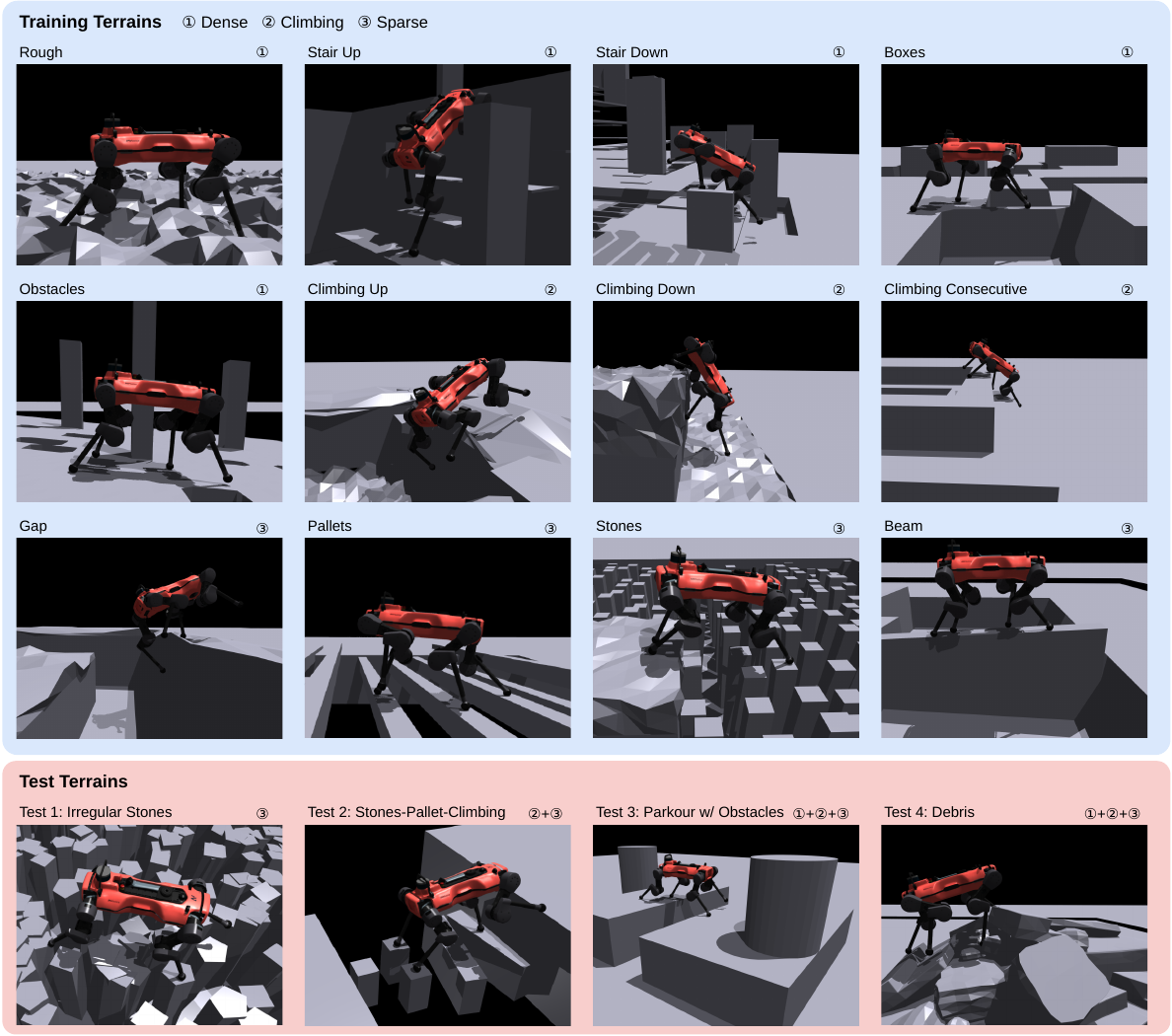}
    \vspace{-0.2cm}
    \caption{Training and Test Terrains. We train our locomotion policies on primitive terrains of three categories: \textcircled{\tiny 1} dense, \textcircled{\tiny 2} climbing, and \textcircled{\tiny 3} sparse. To test generalization, we evaluate the policies on four terrains that are either entirely unseen or combinations of training and/or unseen terrains.}
    \vspace{-0.3cm}
    \label{fig:terrains}
\end{figure*}

We use three types of rewards shared across both quadruped and biped training environments, as listed in \Table{rewards}: task rewards which incentivize goal-reaching behaviors, regularization rewards which improve stability and safety, and simulation-fidelity rewards, which penalize near-limit joint states that can cause unrealistic simulation.

There are four task reward terms, respectively for goal position tracking, goal heading tracking, moving towards goal, and standing at goal. 
Following \cite{he2024agile}, we define the position tracking reward as
\begin{equation}
    r_{\rm position\_tracking} = \frac{1}{1+0.25 d_{xy}^2} \cdot t_{\rm mask}(4),
    \label{eq:rew_pos_tracking}
\end{equation}
where $d_{xy}$ is the horizontal distance from the robot to the goal position, and $t_{\rm mask}(\cdot)$ is a time-based mask function defined as
\begin{equation}
    t_{\rm mask}(T) = \frac{1}{T} \cdot \mathbf{1}(t_{\rm left} < T),
\end{equation}
with $t_{\rm left}$ denoting the remaining time of the current episode. Further, we define the heading tracking reward as
\begin{equation}
    r_{\rm heading\_tracking} = \frac{1}{1+d_{\rm yaw}^2} \cdot t_{\rm mask}(2) \cdot \mathbf{1}(d_{xy}<0.5)
    \label{eq:rew_yaw_tracking},
\end{equation}
Intuitively, these two tracking rewards encourage the robot to be at the goal position with the desired heading at the end of the episode, without constraining how it reaches the goal, thereby allowing complex locomotion skills to emerge for traversing terrains.

To further facilitate exploration, we introduce a moving-to-goal reward:
\begin{equation}
\begin{split}
    r_{\text{move}} = \mathbf{1}\Big( & d_{xy} < 0.5 \;\lor \\
    & \big( \cos \theta_{v_b, \text{goal}} > 0.5 \land v_{\min} \le \| [{v}_b]_{xy} \| \le v_{\max} \big) \Big),
\end{split}
    \label{eq:rew_move}
\end{equation}
where $\theta_{v_b, \text{goal}}$ is the angle between the base velocity and the vector from the base to the goal position, $v_{\min}=0.3\,\text{m/s}$ is the lower bound of the horizontal base velocity $[{v}_b]_{xy}$ to be considered as moving, and $v_{\max}=2\,\text{m/s}$ is an upper bound chosen according to hardware and SLAM constraints.
In other words, $r_{\text{move}} = 1$ if the robot is already close to the goal or if it is moving roughly towards the goal, and $r_{\text{move}} = 0$ otherwise. 

To enforce a stable standing posture after reaching the goal, we define the standing reward:
\begin{equation}
\begin{split}
    r_{\rm stand} = & \mathbf{1}(d_{xy}<0.5 \land d_{\rm yaw}<0.5) \\
    & \cdot \exp \left( - \frac{d_{\rm foot} + d_{\rm g} + d_{\rm q} + d_{xy}}{4} \right),
\end{split}
    \label{eq:rew_stand}
\end{equation}
where $d_{\rm foot}$ is the number of feet not in contact divided by the total number of feet, $d_{\rm g} = 1-[g_b]_z^2$ measures the base tilt relative to gravity, and $d_{\rm q}$ is the mean deviation of joint positions from the standing reference. This reward ensures the robot maintains a static, upright configuration at the goal.

The regularization and simulation fidelity rewards are presented in \Table{rewards}. We set all weights to integer powers of 10, and make each term's contribution $1 \sim 2$ orders of magnitude smaller than the task rewards. This enables successful learning without extensive weight tuning. 

In the regularization rewards, we penalize each occurrence of these undesired events:
\begin{itemize}
    \item Spinning too fast: Yaw rate $\lvert [\omega_b]_z \rvert > 2.0$\,rad/s, which can trigger drifts in the odometry.
    \item Leaping on flat terrain: All feet are off the ground when the elevation difference is $< 30$ cm.
    \item Non-foot contacts: We penalize both general non-foot contacts and non-foot contact switches (from no contact to contact). We separate these to allow stable interactions when necessary (e.g., climbing) while discouraging unnecessary non-foot contacts.
    \item Stumbling: Any link having a horizontal contact force larger than the vertical force.
    \item Slippage: Any link moving while in contact.
    \item Self-collision: Collisions between robot links.
\end{itemize}
Unlike previous works~\cite{wang2025beamdojo, dong2025marg, yu2024walking}, we do not explicitly reward or penalize foot contact positions in sparse terrains; instead, we formulate rewards for all robot links to enable emergent whole-body contacts. Those foothold position rewards are also difficult to define in a terrain-agnostic way: for example, quadruped climbing benefits from active knee contacts and near-edge foot placements, as shown in \fig{result_parkour}, whereas on sparse terrains near-edge foot placements are undesirable.

\medskip 

\subsubsection{Termination}
\label{sec:subsec_termination}
To facilitate training, we set the following early termination conditions:
\begin{itemize}
    \item Bad orientation: The projected gravity vector satisfies $|[g_b]_x| > 0.985$, $|[g_b]_y| > 0.7$, or $[g_b]_z > 0.0$ (robot flipped).
    \item Base collision: The base link experiences a contact force larger than the robot's total weight.
    \item High thigh acceleration: The acceleration of any thigh link exceeds a specific threshold {when the corresponding foot is in contact}. We design this to reduce impact during jumping, as stiff landings can damage hardware. Interestingly, we use typical data from dog ($60\rm\ m/s^2$~\cite{pfau2011kinetics}) and human ($100\rm\ m/s^2$~\cite{mcerlain2021surface}) biokinetics literature, and they directly work to emerge impact absorption or avoidance behaviors, which is shown in \secref{results}.
    \item Stagnation: Movement in the past 5\,s is less than 0.5\,m while the robot is still $> 1$\,m away from the goal position.
\end{itemize}
Triggering any of these conditions terminates the episode immediately and incurs early termination penalties.

\medskip

\subsubsection{Terrains and Curriculum}

We train our controllers on primitive terrains and evaluate generalization on complex test terrains. Both are shown in \fig{terrains}, while quantitative results are presented in \secref{benchmark}. We use a terrain curriculum that scales difficulty from easy to hard (detailed in Appendix~\ref{sec:app_terrain}), similar to \cite{zhang2024risky}. To stabilize learning across different challenging terrains, we estimate the robot's success rate using an exponential moving average. If the robot reaches the goal and its estimated success rate exceeds 0.5, we promote the environment to the next difficulty level. Conversely, the level is demoted if the robot remains $>4$~m away from the goal (which is the average starting distance). Once the robot passes the highest level, the environment resets to a random difficulty.

Additionally, we use a perception noise curriculum and an initial heading curriculum. In the first 20\% iterations of teacher training, we linearly increase mapping noise from zero to the maximum level, and simultaneously expand the initial heading from facing the goal to a random yaw in $[-\pi, \pi]$.

\medskip
\subsubsection{Domain Randomization}


We employ domain randomization~\cite{tobin2017domain} during training to enhance robustness and facilitate sim-to-real transfer. We apply the following randomization setup:
\begin{itemize}
\item {Robot Dynamics:} We uniformly randomize the payload, friction coefficients, and actuation delays. For the TRON1 biped, we additionally randomize PD gains and motor armatures, following~\cite{he2025attention}.
\item {Observation Noise:} We apply uniform noises to the policy observations. We also degrade the student's depth clouds by simulating missing points and sensor artifacts.
\item {Mapping:} During student training, we randomly select a subset of environments to access complete maps, while others rely on partial online maps. This setup enables map reuse when the robot traverses the same terrain repeatedly. We also corrupt the student's mapping by randomly removing points and assigning random height values with high uncertainties. Additionally, we simulate mapping drifts for both the teacher and the student.
\end{itemize}
Implementation details are provided in Appendix~\ref{sec:app_rand}.

\subsection{Training and Deployment}

\subsubsection{Training Setup}

We train our controllers in Isaac Gym~\cite{makoviychuk2021isaac} with the PPO implementation in RSL-RL~\cite{schwarke2025rsl}. The both teacher policies are trained with 80000 iterations, and the both student policies are trained with 40000 iterations (surrogate loss disabled during first 5000 iterations). The hyperparameters are listed in Appendix~\ref{sec:app_ppo}. The training cost for the ANYmal-D policies is $\sim60$ RTX-4090-days with $8$ GPUs in parallel. The training cost for the TRON1 policies is $\sim30$ RTX-4090-days with $4$ GPUs in parallel.

The difference between both robots' training costs are due to the mapping sizes. For ANYmal-D, we use a map of size $36\times 14$, with $8$-cm resolution, centered at $x=0.6 \rm{\ m}, y=0 \rm{\ m}$ in the base frame. For TRON1, we use a map of size $18 \times 13$, with $8$-cm resolution, centered at $x=0.32 \rm{\ m}, y=0 \rm{\ m}$ in the base frame. These numbers are designed based on the robot dimensions and terrain sizes.

\medskip
\subsubsection{Deployment Setup}

We deploy the controllers using ONNX Runtime~\cite{onnxruntime}. On the onboard Intel Core i7-8850H CPUs, the policy inference time is approximately $2$~ms.

\medskip
\subsubsection{Other Sim-to-Real Designs} We model the actuator dynamics for sim-to-real transfer. In simulation, we use the actuator network~\cite{hwangbo2019learning} for ANYmal-D, and an identified DC-motor model for TRON1. We clip the applied torques using the joint torque-velocity constraints in \cite{shin2023actuator}.

To enable continuous deployment (infinite-horizon execution), we define the command observations differently from prior goal-reaching works~\cite{rudin2025parkour, he2024agile, 10611254}. Following prior practice, the critic receives the full command: the goal’s relative position, the sine and cosine of the relative yaw, and the remaining episode time. The actor, however, receives a modified command representation: we clip the observed goal distance to a maximum of $2\,\text{m}$ and remove the remaining time. Additionally, if the actual goal distance exceeds $2\,\text{m}$, we randomize the observed yaw command during training. This design decouples policy deployment from finite-horizon training and simplifies steering control.

\section{Neural Mapping}
\label{sec:mapping}

\subsection{Mapping Pipeline}
\begin{figure*}[ht]
    \centering
    \includegraphics[width=1\linewidth]{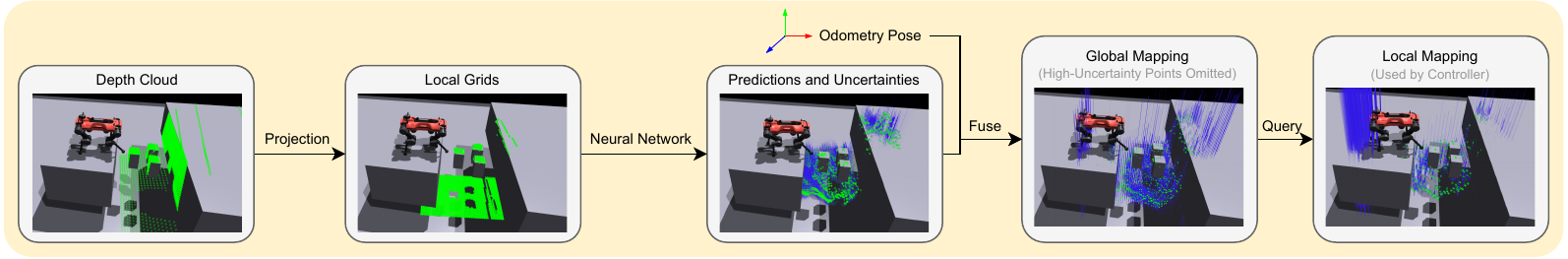}
    \vspace{-0.5cm}
    \caption{Our proposed mapping pipeline. For each frame of depth clouds, we project it into local grids, and use a lightweight neural network to predict elevation estimations with uncertainties. The predictions are then fused into the global map via the odometry, and local maps can be queried from the global map to serve as the controller inputs. In visualization, the green points indicate the depth points or elevation estimations, and the blue lines indicate the uncertainties.}
    \vspace{-0.3cm}
    \label{fig:mappingpipe}
\end{figure*}

Our proposed neural mapping pipeline is illustrated in \fig{mappingpipe}. It can not only achieve real-time computation on the hardware, but also run with thousands of parallel environments in simulation, thereby bridging the sim-to-real gap with identical mapping pipelines.

For each sensing frame, we project the point cloud onto a local 2D height grid. If multiple points fall into the same cell, we keep the maximum $z$-value, which is most relevant for locomotion, and we assign a fixed minimum value to cells with no points. The resulting local elevations, however, are often noisy and incomplete due to occlusions. To mitigate this, we use a lightweight CNN trained with Bayesian learning (detailed in \secref{sub_training}) to jointly predict base-relative elevations and their uncertainties (in the form of log-variance). The predicted uncertainties capture both measurement noise and occlusions, while the elevation predictions themselves can also suppress noise.

\begin{figure}[t]
    \centering
    \includegraphics[width=0.9\linewidth]{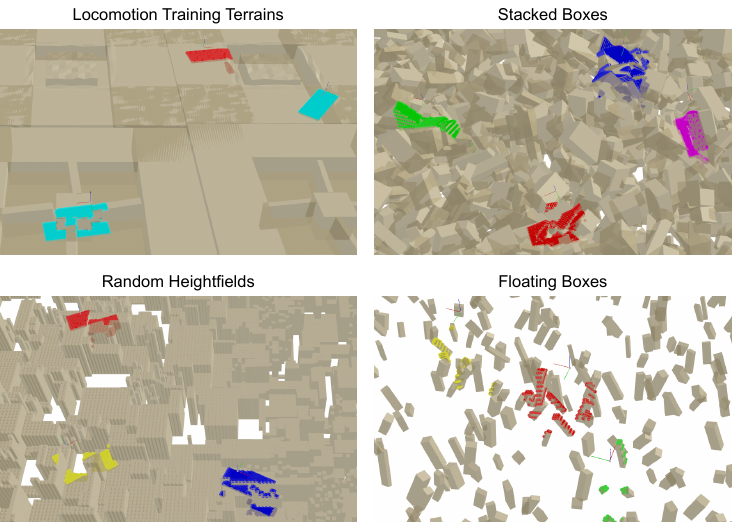}
    \vspace{-0.35cm}
    \caption{Terrain meshes used for mapping model training. We use four meshes: the locomotion training terrains, randomly stacked boxes, random heightfields, and random floating boxes. The colored points visualized are the elevation scans from random poses.}
    \vspace{-0.5cm}
    \label{fig:mappingterrain}
\end{figure}

We then use the odometry poses {(or world-frame poses in simulation)} to fuse these local predictions into a global grid map $\mathcal{M}$. It has two layers: the elevation layer and the uncertainty layer (in the form of variance). The global map is initialized as flat ground at the robot’s standing height (ground height relative to the base) with large uncertainties. At each new frame, given the local elevation estimations with uncertainties and the current base pose, we project the local predictions to the global grid cells. For any global grid point $(u,v)$ covered by the local grids, we denote the new estimation as $h_t$ and its uncertainty as $\sigma_t^2$, fusing them with existing values $h_{prior}$ and $\sigma^2_{prior}$.

We do not use standard Bayesian fusion because repeated observations of the same occluded or uncertain area should not have reduce uncertainties simply due to consistent predictions. Instead, we employ a \textit{Probabilistic Winner-Take-All} strategy. First, we calculate an effective measurement variance $\hat{\sigma}^2_t$ that is lower-bounded by the prior to prevent over-confidence:
\begin{equation}
    \hat{\sigma}^2_t = \max(\sigma^2_t, 0.5 \cdot \sigma^2_{prior}).
\end{equation}
An update is considered valid only if the effective measurement variance is not significantly larger than the prior ($\hat{\sigma}^2_t < 1.5 \sigma^2_{prior}$), or if the absolute uncertainty is low ($\hat{\sigma}^2_t < 0.2^2$).

For valid updates, we determine the probability $p_{\rm win}$ of overwriting the map based on the relative precision:
\begin{equation}
    p_{\rm win} = \frac{(\hat{\sigma}^2_t)^{-1}}{(\hat{\sigma}^2_t)^{-1} + (\sigma^2_{prior})^{-1}}.
\end{equation}
Finally, the map is updated stochastically. We sample $\xi \sim \mathcal{U}[0,1]$ and let the new prediction take over the cell if the sample falls within the probability threshold:
\begin{equation}
    (h_{new}, \sigma^2_{new}) \leftarrow 
    \begin{cases} 
        (h_t, \hat{\sigma}^2_t) & \text{if } \xi < p_{\rm win} , \\
        (h_{prior}, \sigma^2_{prior}) & \text{otherwise.}
    \end{cases}
\end{equation}
For the controller inputs, we then just query the grids around the robot pose in the latest global map.

This \textit{Probabilistic Winner-Take-All} strategy offers the following benefits:
\begin{itemize}
    \item The uncertainty of the same occluded point will not decrease through consistent predictions.
    \item Over-confident predictions, if not consistent, cannot take over the cell.
    \item The system can rapidly update the map in response to dynamic terrain changes when high-confidence measurements are available.
    \item The pipeline is easy to integrate with parallel simulation, and fast enough to run on the hardware.
\end{itemize}

\subsection{Training}
\label{sec:sub_training}

\begin{figure*}[ht]
    \centering
    \includegraphics[width=0.95\linewidth]{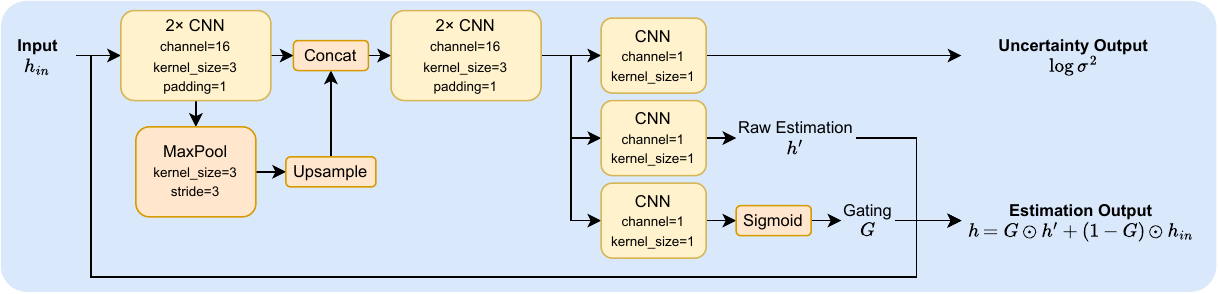}
    \vspace{-0.4cm}
    \caption{Mapping model architecture. We use a lightweight U-Net~\cite{ronneberger2015u} model with a gated residual design for the estimation.}
    \vspace{-0.4cm}
    \label{fig:mappingmodel}
\end{figure*}

We train a CNN to predict per-frame elevation estimates and uncertainties with synthetic data and random terrains. The technical implementations are detailed below.

\subsubsection{Terrains and Data Sampling}

We use both the locomotion training terrain meshes and additional procedurally generated terrains to train the mapping model, as shown in \fig{mappingterrain}. Instead of relying on physical simulation, we directly sample local elevation grids from random poses above these meshes using raytracing with Warp~\cite{warp2022}, enabling sampling at hundreds of thousands of frames per second on a single RTX 4090 GPU.

\subsubsection{Training Data Synthesis}

We apply the following augmentations to the sampled local elevation grids:
\begin{itemize}
    \item additive uniform noise with random magnitude on each cell;
    \item random cropping of the map from the four borders;
    \item simulated occlusions with random sensor {poses} and field of view;
    \item random ranges to clip the elevations;
    \item random missing points and outliers at random ratios.
\end{itemize}
By doing so, we synthesize diverse, noisy, and partially observable local grids as the model inputs, and the original ground-truth elevations as the labels.

\subsubsection{Model and Optimization}

We train the model to reconstruct the ground-truth elevations with $\beta-$NLL loss ($\beta=0.5$) from \cite{seitzer2022pitfalls}:
\begin{equation}
    L_{0.5} = \mathbb{E}_{X,Y} \left[ \mathrm{sg}\left[ \hat{\sigma}(X) \right] \left( \frac{\log \hat{\sigma}^2(X)}{2}  + \frac{(Y - \hat{\mu}(X))^2}{2\hat{\sigma}^2(X)} \right) \right],
    \label{eq:loss}
\end{equation}
where $X$ denotes the inputs, $Y$ denotes the ground-truth elevation {labels}, and $\hat{\mu}(X)$ and $\hat{\sigma}^2(X)$ denote the predicted estimation and variance, respectively. The operator $\mathrm{sg}[\cdot]$ denotes the stop-gradient operation. Compared to the standard negative log-likelihood (NLL) loss used in classical Bayesian learning~\cite{kendall2017uncertainties}, this formulation reduces the tendency of the model to overestimate uncertainty on hard samples to trivially reduce the loss. It encourages the model to output high uncertainty when accurate predictions are not possible, and low uncertainty together with accurate predictions when they are, thereby capturing the noise and occlusions. 

For batched optimization, terrain roughness can vary significantly across samples. As a result, flat terrains can dominate the batch loss and reduce the effective emphasis on challenging cases. To mitigate this, we re-weigh samples in each batch during training using their total variation (TV)~\cite{chambolle2004algorithm}:
\begin{equation}
\label{eq:tv_weight}
\begin{aligned}
\mathrm{TV}(Y_b) &= \frac{1}{HW}\Big(\|\nabla_x Y_b\|_{1}+\|\nabla_y Y_b\|_{1}\Big),\\
w_b &= \frac{\mathrm{TV}(Y_b)}{\sum_{b'=1}^{B}\mathrm{TV}(Y_{b'})+\varepsilon}.
\end{aligned}
\end{equation}
Here, $Y_b$ is the ground-truth elevations of the $b$-th sample in a batch, and $H$ and $W$ are the height and width. The weight $w_b$ is normalized across the batch, and $\varepsilon$ is a small positive constant. By doing so, we assign higher weights to samples with larger elevation variations.

We use a shallow U-Net~\cite{ronneberger2015u} model with a gated residual design, as illustrated in \fig{mappingmodel}. The CNNs output the uncertainty, a raw estimation, and a gating map. The final estimation is obtained by the gated combination of the raw estimation and the input, preserving accuracy in clearly observed areas while selectively overwriting noisy or occluded areas.

We train the models on 54 million frames for each robot, each model takes less than 1 hour to converge. For ANYmal-D, the local grids are of shape $51 \times 31$, with $4$-cm resolution, centered at $x=1.0 \rm{\ m}, y=0 \rm{\ m}$ in the base frame. For TRON1, the local grids are of shape $31 \times 31$, with $4$-cm resolution, centered at $x=0.6 \rm{\ m}, y=0 \rm{\ m}$ in the base frame.

\subsection{Simulation Integration and Deployment}

When deployed in simulation with 1000 parallel ANYmal-D environments, the model’s inference time is below $0.3$~ms, and the GPU memory consumption is about $3$~GB (for TRON1, these numbers are roughly 60\% of those for ANYmal-D). Storing $8,\mathrm{m} \times 8,\mathrm{m}$ global maps for 1000 environments requires about $0.3$~GB, and the global map size can be flexibly chosen since we recenter the map around the robot when it approaches the boundary. Obtaining depth clouds and handling other intermediate overheads takes an additional $\sim 1.2$~GB of GPU memory per 1000 ANYmal-D environments.

On real hardware, the mapping pipeline takes approximately $5$~ms per frame on the onboard CPU, with around $2.5$~ms spent on model inference using ONNX Runtime. This enables fast, low-latency mapping that can keep up with the depth camera’s frame rate. For ANYmal-D, we merge the point clouds from two front facing cameras for processing. For TRON1, we use the only front facing camera.

Regarding odometry, we use CompSLAM~\cite{khattak2025compslam} with Graph-MSF~\cite{nubert2022graph} on ANYmal-D to obtain a high-frequency, accurate LiDAR–inertial odometry. On TRON1, we use DLIO~\cite{chen2023direct} instead, as it {is lightweight and }better handles the platform’s higher accelerations.

Notably, although these odometry solutions can run reliably at high frequency, their velocity estimates are observed to be noisy and delayed. This is intuitive: for example, a 1~cm position drift over a 20~ms timestep can translate to a velocity error of 0.5~m/s, yet has little impact on map fusion. Therefore, we remove the linear velocity observations from the student.

\section{Results}
\label{sec:results}

\subsection{Comparison with Prior Art}

\begin{table*}[!t]
\centering
{
\caption{Comparison with Prior State of the Art (SOTA)}
\label{tab:compare_art}

\resizebox{\textwidth}{!}{
\begin{tabular}{lccccccc}
\toprule
\multirow{2}{*}{Result-wise} & \multicolumn{2}{c}{Quad.\ Climb} & \multicolumn{2}{c}{Biped Climb} & \multirow{2}{*}{Sparse Terrain} & \multicolumn{2}{c}{Generalization} \\
\cmidrule(lr){2-3} \cmidrule(lr){4-5} \cmidrule(lr){7-8}
& Up & Down & Up & Down & & To Debris & To Parkour Line \\
\midrule
SOTA generalization~\cite{he2025attention} & $\sim$0.5\,m$^{A}$ & $\sim$0.5\,m$^{A}$ & $\sim$0.3\,m$^{GR}$ & $\sim$0.3\,m$^{GR}$ & Yes & \textbf{Zero Shot} & No \\
SOTA agility within quadruped generalists~\cite{rudin2025parkour} & \textbf{1\,m}$^{A}$ & \textbf{1\,m}$^{A}$ & -- & -- & Yes & Few Shot & Few Shot \\
SOTA agility within biped generalists~\cite{long2025learning} & -- & -- & 0.5\,m$^{H}$ & $\geq$0.5\,m$^{H,*}$ & No & No & No \\
\textbf{Ours} & \textbf{1\,m}$^{A}$ & \textbf{1\,m}$^{A}$ & \textbf{0.48\,m$^{T}$, 0.65\,m$^{H}$} & \textbf{0.88\,m$^{T}$, 1\,m$^{H}$} & \textbf{Yes} & \textbf{Zero Shot} & \textbf{Zero Shot} \\
\bottomrule
\end{tabular}}

\vspace{2pt}

\resizebox{\textwidth}{!}{
\begin{tabular}{lcccccc}
\toprule
\multirow{2}{*}{Paradigm-wise} & \multirow{2}{*}{Universality: Quad./Biped} & \multirow{2}{*}{Formulation} & \multicolumn{3}{c}{Perception} & \multirow{2}{*}{\shortstack{\#. Stages \\ \revise{or Policies}}} \\
\cmidrule(lr){4-6}
& & & Training & vs. & Deployment & \\
\midrule
SOTA generalization~\cite{he2025attention} & separate rewards & velocity tracking & privileged mapping & $\neq$ & classical mapping & 2 \\
SOTA agility within quadruped generalists~\cite{rudin2025parkour} & quadruped only & \textbf{goal reaching} & \textbf{depth camera \revise{only}} & $=$ & \textbf{depth camera \revise{only}} & $>10$ \\
SOTA agility within biped generalists~\cite{long2025learning} & biped only & velocity tracking & privileged mapping & $\neq$ & classical mapping & \textbf{1} \\
\textbf{Ours} & \textbf{unified rewards} & \textbf{goal reaching} & \textbf{neural mapping} & $=$ & \textbf{neural mapping} & 2 \\
\bottomrule
\end{tabular}}

\vspace{1pt}
{\fontsize{7}{7}\selectfont \hspace{0.1cm } Robot Platforms: $^{A}$ ANYmal-D, $^{GR}$ GR-1, $^{T}$ TRON1, $^{H}$ H1. \hspace{1cm} $^{*}$ Estimated from videos, not reported. \hfill}
}
\vspace{-1mm}
\end{table*}

\begin{figure}[t]
    \centering
    \includegraphics[width=.93\linewidth]{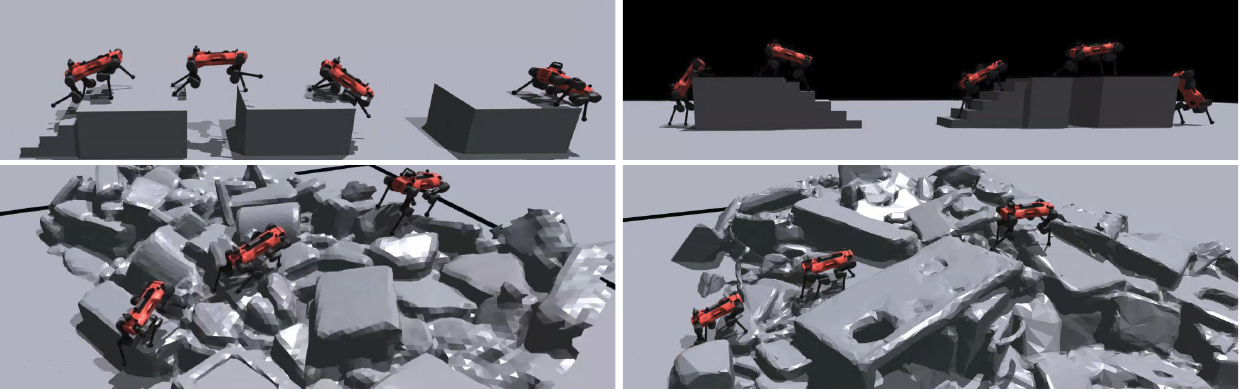}
    \vspace{-3.5mm}
    \caption{Our ANYmal-D controller zero-shots the hardest parkour and rubble pile terrains reported in prior work~\cite{hoeller2024anymal, rudin2025parkour}.}
    \vspace{-5mm}
    \label{fig:simparkour}
\end{figure}

We summarize the comparison with prior work in \Table{compare_art}. Our controllers achieve agility on par with or better than previous state-of-the-art methods, as measured by the obstacle heights they can climb up and down. 
Since comparing TRON1 with the larger and stronger Unitree H1 in~\cite{long2025learning} may be unfair, we also apply our framework to H1 and report the simulation results here and in the videos. \revise{We use our student policies for comparison and deployment.}

On ANYmal-D, our policy is trained only on primitive terrains yet zero-shots the hardest parkour and rubble-pile terrains reported in~\cite{rudin2025parkour} (which required finetuning), as shown in \fig{simparkour}. Our controllers also exhibit stronger generalization than existing methods, as benchmarked in \secref{benchmark}.

{
Besides, our framework also uses unified rewards across embodiments and bridges simulation and real-world perception with the same mapping module, without tuning a real-world mapping pipeline to match the simulation.
}



\subsection{Real World Results}
\subsubsection{Quadruped Parkour}
\begin{figure*}
    \centering
    \includegraphics[width=1\linewidth]{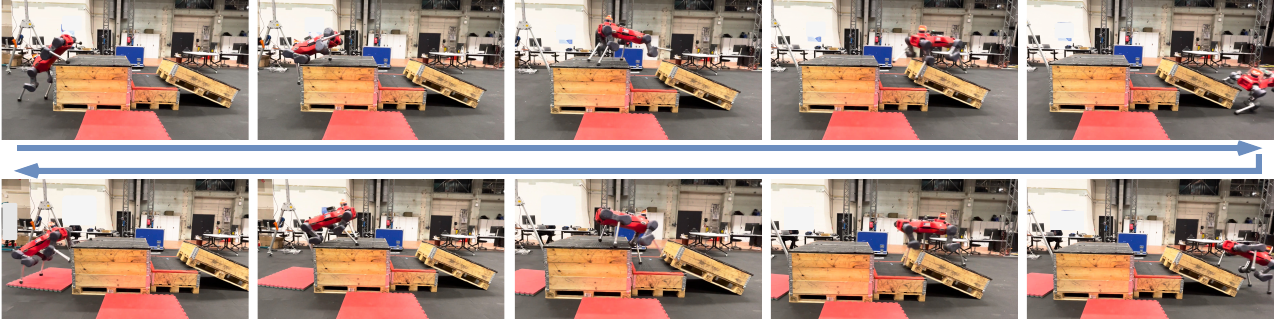}
    \vspace{-0.5cm}
    \caption{Our system enables ANYmal-D to move back and forth on a parkour course, a terrain that is unseen during training of both the controller and the mapping model.}
    \vspace{-0.2cm}
    \label{fig:result_parkour}
\end{figure*}

We demonstrate the agility and generalization of our trained ANYmal-D controller on a parkour course, as shown in \fig{result_parkour}. This course is not included in the training terrains of either the locomotion controller or the mapping model, yet our system can stably traverse it at speeds of up to $2$~m/s, composing climbing and jumping maneuvers.

\subsubsection{Sparse Terrain Locomotion}
\begin{figure*}
    \centering
    \includegraphics[width=1\linewidth]{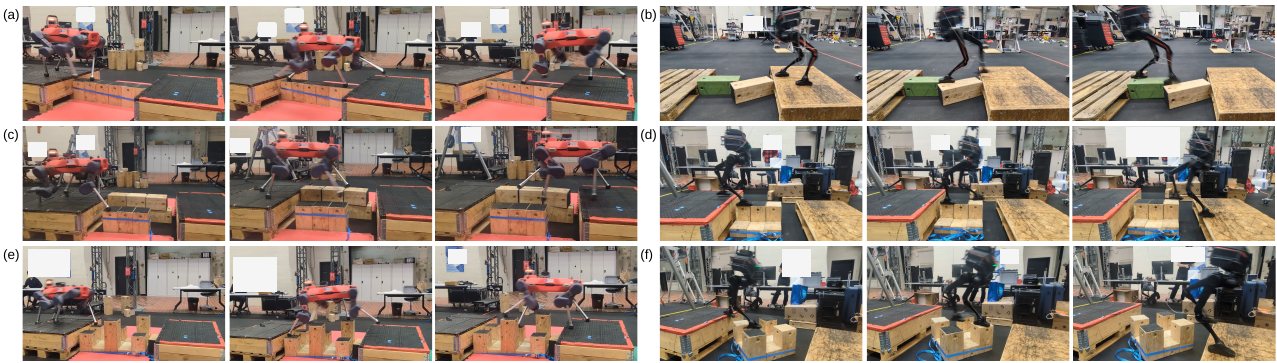}
    \vspace{-0.5cm}
    \caption{Our systems enable ANYmal-D and TRON1 to traverse diverse sparse terrains. (a) ANYmal-D traverses a $19$-cm wide balance beam, which is in the training terrains. (b) TRON1 traverses two unfixed floating $19$-cm wide blocks, which make an unseen curved beam terrain. (c) ANYmal-D traverses two unfixed beams with gaps, which is unseen during training. (d) TRON1 traverses a beam followed by a gap, which is an unseen combination during training. (e) ANYmal-D traverses two rows of $19$-cm wide stepping stones with $20$-cm height differences, which is unseen during training. (f) TRON1 traverses diamond-layout stepping stones, which is unseen during training.}
    \vspace{-0.5cm}
    \label{fig:result_sparse}
\end{figure*}

As shown in \fig{result_sparse}, our controllers enable both ANYmal-D and TRON1 to traverse a variety of sparse terrains, many of which are unseen during training, exhibiting generalization capabilities.

\subsubsection{Biped maneuvering}
\begin{figure}
    \centering
    \includegraphics[width=0.92\linewidth]{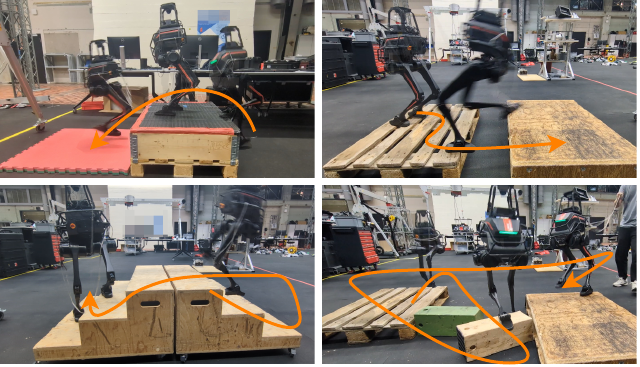}
    \vspace{-0.35cm}
    \caption{TRON1 maneuvers over a 38-cm-high platform, a gap, a staircase, and rough terrain, showcasing its omnidirectional perceptive locomotion capabilities. Orange arrows indicate the robot’s approximate trajectories.}
    \vspace{-0.45cm}
    \label{fig:result_bipedrough}
\end{figure}

To demonstrate the omnidirectional perceptive locomotion capabilities of our controller, we command TRON1 to traverse a 38-cm-high platform, a gap, a staircase, and rough terrain, as shown in \fig{result_bipedrough}. The robot executes smooth maneuvers in all directions.

\subsubsection{Robustness}

\begin{figure}[t]
    \centering
    \includegraphics[width=.9\linewidth]{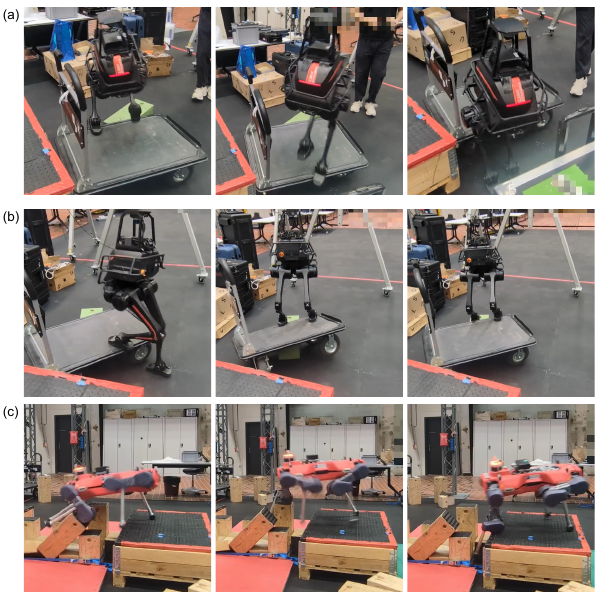}
    \vspace{-0.2cm}
    \caption{Our controllers are robust to moving terrains. (a) TRON1 climbs over an unlocked tiltable platform cart. (b) TRON1 climbs onto and balances on an unlocked tiltable platform cart. (c) ANYmal-D recovers from unfixed tilted stepping stones using its knees.}
    \label{fig:result_robust}
    \vspace{-0.5cm}
\end{figure}

We demonstrate the robustness of our controllers on moving terrains. As shown in \fig{result_robust}, TRON1 can traverse or balance on a platform cart with unlocked wheels and spring suspension that can easily tilt to one side with the robot weight. We also show that ANYmal-D can recover from unfixed tilted stepping stones using its knees, after an occasional trip causes a misstep that tilts the blocks.

\subsubsection{Mapping Results}

\begin{figure}[t]
    \centering
    \includegraphics[width=0.9\linewidth]{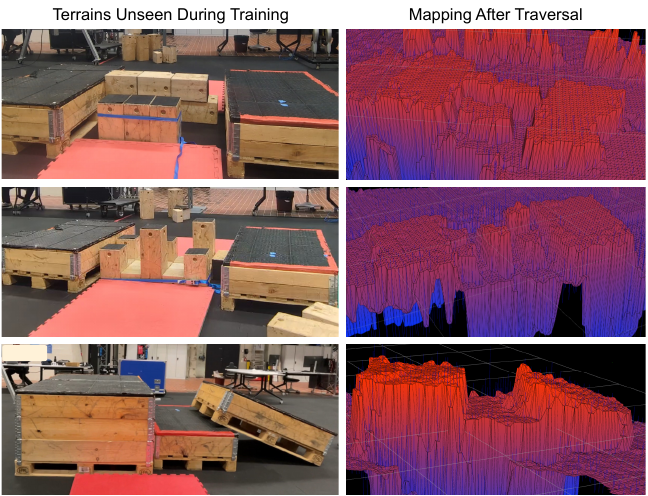}
    \vspace{-0.35cm}
    \caption{Obtained maps after ANYmal-D’s traversal of terrains that are unseen during training. Our mapping supports agile and generalized locomotion. In the visualizations, meshes represent the estimated elevations, and thin lines indicate the uncertainties. The meshes are colored with a red–blue colormap to enhance height contrast, excluding regions that are never covered by the cameras.}
    \vspace{-0.3cm}
    \label{fig:result_mapping}
\end{figure}

In \fig{result_mapping}, we show the mapping results after ANYmal-D traverses terrains that are unseen during training. These maps capture fine-grained terrain details, such as gaps and supports, that help the controller achieve agile and generalized locomotion. Among prior works, only \cite{hoeller2024anymal} has demonstrated maps of comparable quality during agile motions. However, that approach cannot reliably infer unseen parts of obstacles in unstructured terrains~\cite{rudin2025parkour}, is not efficient enough to support concurrent locomotion training, and does not explicitly encode uncertainty under partial observations such as occlusions. For generalized locomotion, explicitly modeling uncertainty is important: occluded regions are not completed by learned priors but kept uncertain, and newly observed geometry can be integrated into the map based on the predicted uncertainty once it becomes visible.

\subsection{Emergent Behaviors}

\subsubsection{Whole-Body Contact}

As previously shown in \fig{result_parkour} and \fig{result_robust}, our ANYmal-D controller actively use knee contacts to help stabilize the robot and traverse challenging terrains. On one hand, such whole-body contact behaviors are beneficial for agility and robustness; on the other hand, because most existing legged robots are designed primarily for foot contacts, these behaviors may stress the hardware. That said, we believe that whole-body contact will become increasingly important for achieving whole-body dexterity in complex environments~\cite{zhang2025wococo, zhuang2025embrace, yang2025omniretarget}.

\subsubsection{Impact Reduction}

\begin{figure}[t]
    \centering
    \includegraphics[width=0.95\linewidth]{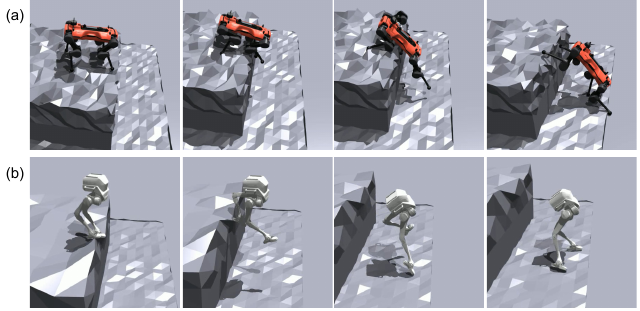}
    \vspace{-0.4cm}
    \caption{Emergent impact reduction behaviors. (a) ANYmal-D uses its knee to support the body when climbing down, and gently touches the ground to reduce the impact. (b) TRON1 extends one leg while jumping down, and retracts it during landing before switching support to the other leg.}
    \vspace{-0.4cm}
    \label{fig:emergent_impact}
\end{figure}

As mentioned in \secref{subsec_termination}, we use human and dog motion data to set thigh-acceleration thresholds for early termination. This induces impact-reduction behaviors, as shown in \fig{emergent_impact}: when climbing down, ANYmal-D uses its knees to gently touch down, while TRON1 retracts the support leg to absorb the impact.

\subsection{Interpretable Feature Patterns}

\begin{figure*}[ht]
    \centering
    \includegraphics[width=0.95\linewidth]{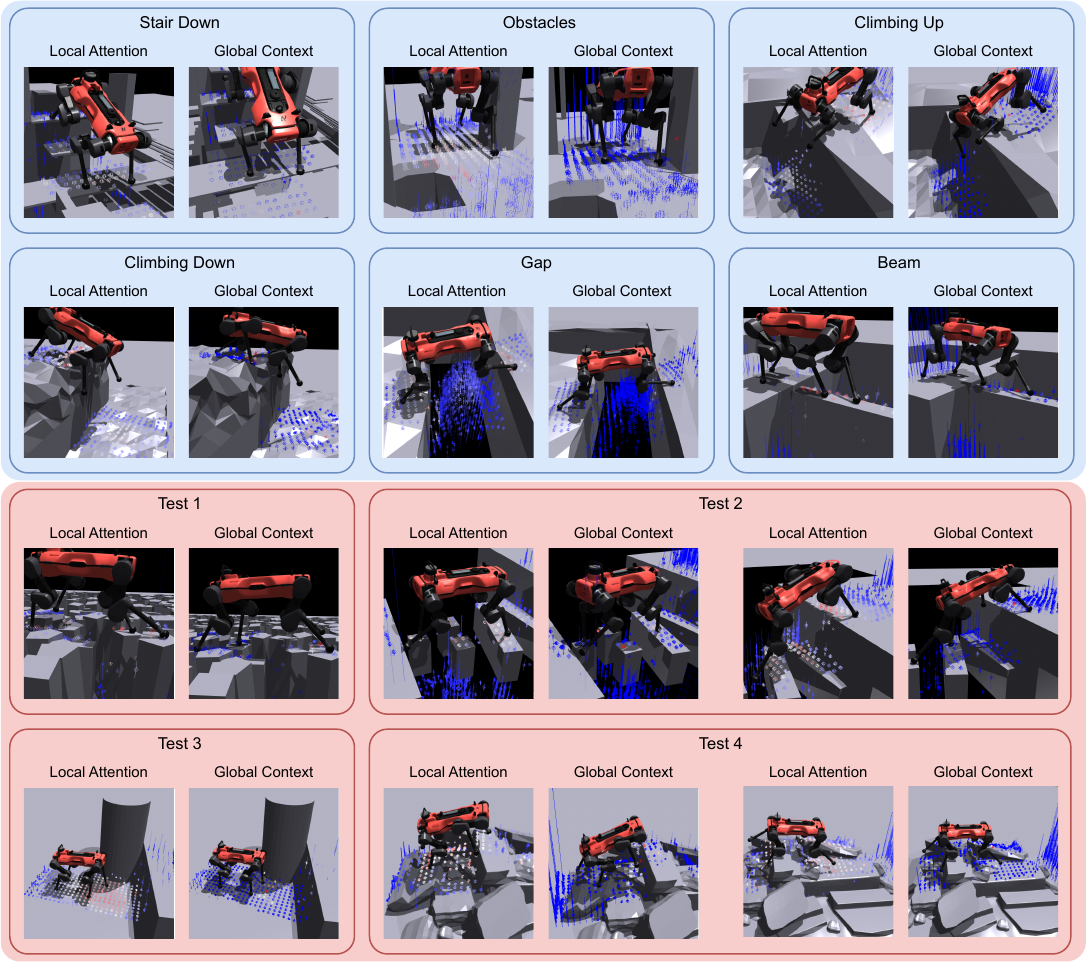}
    \vspace{-0.35cm}
    \caption{Feature patterns of our ANYmal-D controller. We visualize the neural elevation map together with the corresponding local-attention and global-context weights. Points indicate elevation estimates, lines indicate uncertainties, and higher-intensity red colors correspond to higher weights. Local attention weights provide fine-grained patterns for locomotion control, while global context features primarily focus on sparse points that characterize the terrain type.}   
    \vspace{-0.35cm}
    \label{fig:emergent_attn}
\end{figure*}

We visualize the feature patterns of our ANYmal-D student policy's AME-2 encoder in \fig{emergent_attn}. For local features, we visualize their attention weights with a red–blue colormap. For global features, which are obtained via max pooling, we treat pooled points as selected (weight 1) or not (weight 0) and average this binary mask over the feature dimension.

{
By inspecting these feature maps, we observe a consistent empirical pattern: local attention tends to concentrate around future contact regions, whereas the global context focuses on a sparse set of characteristic points that capture terrain-level structure and influence the motion style. This pattern appears on both training and test terrains, suggesting that the learned representation transfers beyond the training distribution.
}

{
On training terrains, the attention patterns vary with terrain type. During stair descent, local attention focuses on steppable regions, while the global context attends to the next stair levels. For obstacle avoidance, local attention focuses on future footholds, while the global context highlights obstacles and free space. During climbing up, local attention covers front-foot landing and hind-leg knee-support regions, while the global context indicates the upcoming climb. During climbing down, local attention covers hind-foot support, front-knee support, and front-foot landing regions, while the global context captures sparser climbing-related cues. For gap and beam traversal, local attention focuses on future foothold or contact regions, while the global context captures sparse support structures that guide forward progression.
}

{
Similar patterns persist on test terrains. On Test 1, the features resemble those on sparse training terrains. On Test 2, they transition from sparse-terrain traversal to climbing: global-context points move to the platform top, while local attention shifts from future footholds to knee-support regions. On Test 3, the features first resemble obstacle-avoidance patterns before reacting to the climbing terrain. On Test 4, local attention highlights feasible foot and knee contacts on the debris mesh, while the global context captures whether the robot is climbing up or down. These results suggest that the encoder reuses and recombines learned contact-reasoning patterns on unseen terrains.
}

{
We note that local attention does not always concentrate on an exact point and can be smooth, especially on nearly flat contact regions. This is expected because the features are not explicitly supervised with contact labels, and such smoothness does not necessarily degrade the embedding. Enforcing a stronger correspondence between attention and contact patterns remains an interesting direction for future work.
}

\section{Ablations and Benchmarking}
In this section, we benchmark our design choices on test terrains (depicted in \fig{terrains}) that are unseen during training. All benchmarking experiments are conducted on ANYmal-D in simulation, where richer prior works and baselines are available.

\label{sec:benchmark}

\subsection{Teacher Architecture}


\begin{figure*}[ht]
    \centering
    \includegraphics[width=0.98\linewidth]{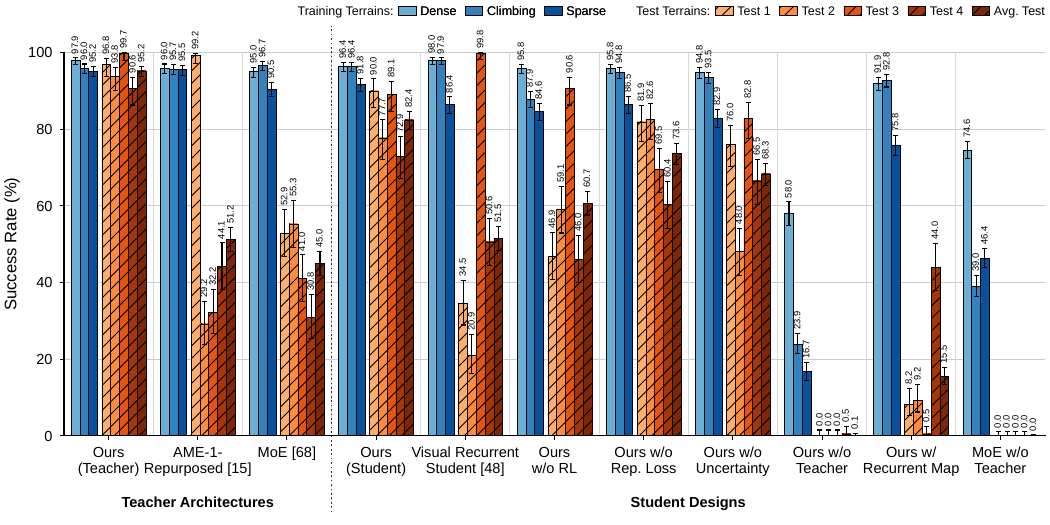}
    \vspace{-0.3cm}
    \caption{\revise{Success Rates (\%) of Different Teacher Architectures and Student Designs on Training and Test Terrains. 0.95 Wilson confidence intervals visualized.}}
    \vspace{-0.3cm}
    \label{fig:ts_succrate}
\end{figure*}

We compare our teacher policy architecture against two baselines: {repurposed} AME-1~\cite{he2025attention} and MoE~\cite{chen2025gmt}. AME-1 is the prior state of the art in generalized locomotion, outperforming policies based on MLPs, CNNs, multimodal transformers~\cite{yanglearning}, and Vision Transformers~\cite{dosovitskiy2020vit}. {It is repurposed and trained under our environment setup to provide a controlled comparison on generalization.} The MoE architecture from~\cite{chen2025gmt} has demonstrated strong scalability by enabling a humanoid to track diverse human motions.

The results are summarized in \revise{\fig{ts_succrate}'s left part}. All architecture designs scale well on the training terrains, highlighting the stability of our training framework. However, MoE shows very limited generalization to unseen test terrains. AME-1 generalizes well on sparse terrains (test 1) but struggles when different terrains are mixed (test 2-4), which we attribute to its encoder assigning attention to local features based only on proprioception, thus missing global context. In contrast, our \gst{AME-2} teacher generalizes reliably across all test terrains.

\subsection{Student Designs}

Having a generalized teacher does not automatically yield a generalized deployable student. We therefore compare our \gst{AME-2} student against the following baselines:
\begin{enumerate}
\item The end-to-end student design from~\cite{rudin2025parkour}, the previous state of the art in agile generalist locomotion, which uses recurrent networks with vision inputs.
\item \revise{Our} student trained without RL losses, using only teacher action distillation and the representation loss.
\item \revise{Our} student trained without the representation loss.
\item {\revise{Our} student trained without uncertainty inputs.}
\item {\revise{Our} student trained from scratch without teacher supervision, under the same training budget.}
\item {\revise{Our} student trained with a temporal recurrent mapping model (will be introduced in \secref{subsec_badmaps}).}
\item \revise{MoE student trained from scratch without teacher supervision.}
\end{enumerate}
The results are summarized in \revise{\fig{ts_succrate}'s right part}. On the training terrains, all student policies perform well, with our proposed design achieving slightly higher success rates on sparse terrains. On unseen test terrains, our student achieves the best overall success rates.

The end-to-end visual recurrent student from~\cite{rudin2025parkour} performs slightly better than ours on \textit{Test 3}, a parkour course with obstacles, but underperforms on all other test terrains that contain sparse regions. On \textit{Test 3}, most of our student’s failures occur when it attempts to climb the obstacles, which can be misinterpreted as higher platforms due to occlusions and the observed surface elevations. In contrast, using the full depth image, as in the visual recurrent design, may provide richer cues and a longer reaction horizon for obstacle avoidance. Despite this modest drop in generalization around obstacles, our student controller achieves much higher success rates on sparse and mixed terrains.

Regarding the learning setup, we find that direct teacher supervision (action distillation plus representation alignment) already yields good performance on the training terrains, but does not produce strong generalization to unseen terrains without RL. Incorporating RL is important for performance on the test terrains, and the representation loss gives a further improvement. {Using only RL without teacher supervision, on the other hand, makes it hard for the student to explore locomotion skills \revise{for both our architecture and MoE}.} Overall, our student policy architecture performs better on unseen terrains than the end-to-end visual recurrent policy, and our learning design, with mixed RL losses and teacher supervision losses (action distillation and representation alignment), further enhances the generalization. {By comparing against an uncertainty-free student baseline, we further show that mapping uncertainty provides useful information for locomotion.}

\subsection{Mapping Designs}

\begin{table}[t]
\centering
\caption{$L_{0.5}$ Loss of Different Neural Mapping Methods on Training and Test Terrains}
\label{tab:mapping}
\resizebox{\columnwidth}{!}{%
\begin{tabular}{clccclccccc}
\hline
\multirow{2}{*}{Mapping Pipeline} &  & \multicolumn{3}{c}{Training Terrains} &  & \multicolumn{5}{c}{Test Terrains} \\ \cline{3-5} \cline{7-11} 
      &  & Dense & Climbing & Sparse &  & Test 1 & Test 2 & Test 3 & Test 4 & avg. Test \\ \hline
\textbf{Ours}            & & -0.006 & -0.108 & 0.020 & & \textbf{0.033} & 0.227 & \textbf{0.036} & -0.111 & \textbf{0.046} \\
Ours (Loco-Only) & & 0.020   & -0.090 & 0.081 & & 0.104 & 0.234 & 0.087 & -0.075 & 0.088 \\
Temporal Recurrent       & & \textbf{-0.162} & \textbf{-0.114} & \textbf{-0.101} & & 0.316 & \textbf{0.135} & 0.066 & \textbf{-0.176} & 0.085 \\
\hline
\multicolumn{11}{l}{\footnotesize Best (lowest) results are shown in \textbf{bold}.} \\
\end{tabular}}\vspace{-0.4cm}
\end{table}

\begin{figure*}[ht]
    \centering
    \includegraphics[width=.93\linewidth]{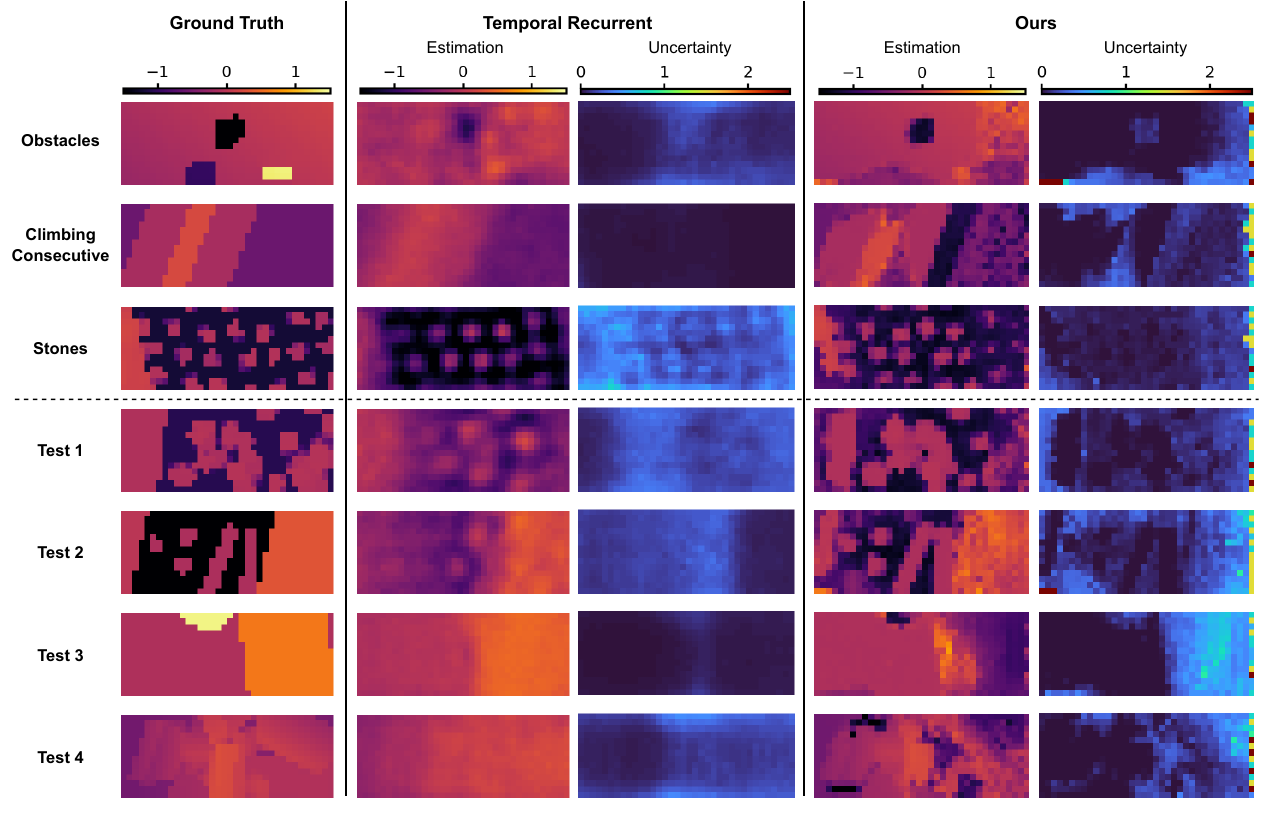}
    \vspace{-0.3cm}    
    \caption{Qualitative comparison between the temporal recurrent model and our mapping pipeline on training and test terrains. The visualized maps are egocentric, with the robot facing rightward. Our mapping pipeline produces accurate estimations when confident and assigns high uncertainty to occluded regions. In contrast, the temporal recurrent model generates less meaningful uncertainty maps, loses fine-grained details, and performs worse on unseen terrains. }
    \vspace{-0.3cm}
    \label{fig:mapping_compare}
\end{figure*}

We benchmark our mapping pipeline against two baselines:
\begin{enumerate}
\item Our pipeline, with the mapping model trained only on locomotion training terrains.
\item A temporal recurrent model that directly predicts egocentric elevation maps from local grid observations and delta transforms.
\end{enumerate}
The first baseline tests whether additional diverse meshes (both traversable and untraversable) are necessary for better generalization. The second contrasts our per-frame fusion approach with an end-to-end recurrent neural memory model: the recurrent model tends to overfit terrain patterns seen during training and produces lower-quality maps on unseen terrains, as we will show in the qualitative comparison in \fig{mapping_compare}.

All pipelines are evaluated on data collected from the same rollout trajectories, with results reported in \Table{mapping}. We use the $L_{0.5}$ loss defined in Eq.~\ref{eq:loss} as the metric, which balances estimation accuracy and uncertainty quantification. The temporal recurrent model (detailed in Appendix~\ref{sec:t_rec_map}) uses a CNN encoder for local grids, an MLP encoder for delta transforms, and Spatially-Enhanced Recurrent Units (SRU), an architecture that enables 100~m-scale mapless visual navigation~\cite{yang2025spatially}, as the memory module. To stabilize training and prevent the model from collapsing to large uncertainties and poor estimates, we have to train the recurrent model only on locomotion training terrains with the policy in the loop (as done in prior work~\cite{yu2024walking, sun2025dpl}) and add an auxiliary L2 loss on the elevation estimates in addition to $L_{0.5}$. This setup requires roughly $5\times$ more training time to converge than our mapping models.

It is worth noting that the recurrent model can optimize both the elevation estimates and the associated uncertainties in unseen regions based on the training distribution, which is advantageous under the $L_{0.5}$ loss, whereas our pipeline uses heuristic defaults. However, as illustrated in \fig{mapping_compare}, our mapping pipeline produces higher-quality maps in practice, and can effectively model occlusions with high uncertainties.

\subsection{Robustness to Visual Noise}

\begin{figure}[t]
    \centering
    \includegraphics[width=0.98\linewidth]{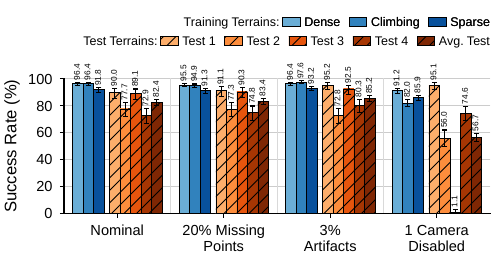}
    \vspace{-0.3cm}
    \caption{\revise{Success Rates (\%) of the Student Policy under Noise and Sensor Degradation. 0.95 Wilson confidence intervals visualized.}}
    \vspace{-0.5cm}
    \label{fig:S-noise}
\end{figure}

We validate the robustness of our system to noise and sensor degradation. We evaluate our student policy’s success rates under missing points (removed from depth clouds), artifacts (replaced by random points in depth clouds), and with the front upper camera disabled. The tested setups are all above the randomization ranges during training.

The results are presented in \revise{\fig{S-noise}}. Under missing points and artifacts, we find the policy performance does not drop, and can sometimes be slightly better. This can be explained by that, fewer effective points in the depth clouds result in more uncertain regions and make the policy behave more conservatively. 

When the front upper camera is disabled, the system still performs well on most terrains without re-training, highlighting the flexibility of our method for different sensor configurations. However, on terrains that require {interactive} perception, such as the high obstacles to climb in \textit{Test 2} and \textit{Test 3}, the policy struggles, suggesting that these behaviors remain sensitive to the available sensor viewpoints.

\subsection{{Limitations of Mapping Baselines for Locomotion}}
\label{sec:subsec_badmaps}

\begin{figure}[t]
    \centering
    \includegraphics[width=0.8\linewidth]{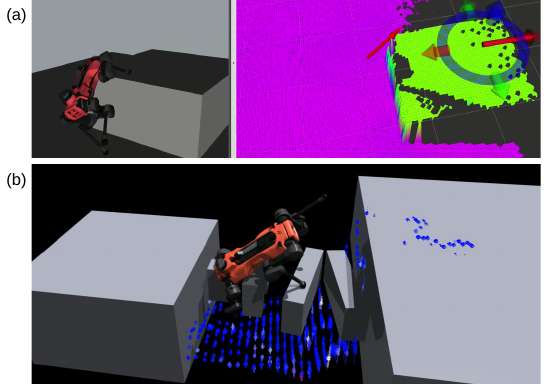}
    \vspace{-0.3cm}    
    \caption{{Failures of applying mapping baselines for locomotion. (a) The classical mapping pipeline~\cite{takamapping} produces occlusion-induced missing values, which degrade locomotion. Gazebo simulation is shown on the left; the mapping, robot pose, and goal marker are shown on the right. (b) The temporal recurrent mapping model observes the high platform in the Test 2 terrain and incorrectly predicts flat ground under the feet.}}
    \vspace{-0.3cm}
    \label{fig:badmapping}
\end{figure}

{We further evaluate perception-in-the-loop locomotion with baseline mapping methods and show that mapping quality can bottleneck locomotion performance. In \fig{badmapping}, we show common failure cases of the classical mapping pipeline~\cite{takamapping} used in~\cite{he2025attention} and the temporal recurrent mapping model.}

{
For the classical pipeline, we deploy a teacher policy trained with the same mapping-noise setup as in~\cite{he2025attention} in the Gazebo digital twin. During evaluation, we remove sensor noise, use ground-truth state estimation, neglect computational limits, and apply neighborhood minimum filters to missing values (which ensures gaps in sparse terrains are not filled). Nevertheless, the policy fails to climb onto a box due to occlusion-induced missing values that are not modeled during training. Alternative hand-crafted filters may address this specific case, but designing filters that generalize across diverse terrains is impractical. 
}

{
For the temporal recurrent mapping baseline, we train another student with it (quantitative results in \fig{ts_succrate}) and find limited mapping generalization also limits locomotion: on the high-platform terrain in Test 2, the mapper incorrectly predicts a dense floor under the feet, causing the robot to misstep and fail to recover. This likely reflects overfitting to climb-up training terrains with near-flat floors in front. Targeted terrain randomization may alleviate this failure mode, but covering all real-world terrain geometries in simulation remains impractical.}

\section{Discussion}

In this paper, we propose \gst{AME-2,} a unified RL framework for agile and generalizable legged locomotion with attention-based neural map encoding. We use goal-reaching rewards to incentivize agile behaviors, and design the AME-2 encoder to achieve generalization across diverse challenging terrains. To facilitate sim-to-real deployment, we combine teacher-student learning with RL and introduce a lightweight neural mapping pipeline that explicitly models partial observability while running identically in simulation and on real robots. Our framework applies to both {quadrupedal and bipedal robots} without changing the rewards or training settings.

\subsection{{Justification of Global Features in the Encoder}}

{
The global feature in our enhanced attention-based encoder is motivated by contact reasoning beyond local foothold selection. The original AME-1 encoder in~\cite{he2025attention} uses a vanilla attention module analogous to model-based foothold selection, where candidate foot contacts are scored from local terrain features and the robot state. This local formulation is insufficient for complex terrains: high-obstacle climbing may require whole-body contacts, while collision-avoidance scenarios require avoiding undesirable contacts with obstacles such as pillars. Thus, the policy must reason not only about where to step, but also about which contacts to make or avoid.
}

{
In this setting, relying only on local features can encode terrain-specific patterns that exploit the training terrains but fail on coupled climbing and precise-contact scenarios, as shown in Test 2/3/4. We therefore introduce a global feature to modulate local features, enabling contact-pattern selection based on the overall terrain context. Our visualizations show that this global-context reasoning transfers to unseen terrains and improves generalization.
}

\subsection{Modular v.s. End-to-End}

Classical locomotion systems~\cite{jenelten2022tamols, grandia2023perceptive} use modular designs (perception, planning, and control) to make the problem tractable, and can generalize well when their underlying assumptions hold (e.g., low uncertainty, stable foot contacts, accurate mapping and dynamics modeling). Recent works instead reduce system complexity through end-to-end sensor-to-motor policy learning~\cite{rudin2025parkour, zhuang2025humanoid}, but have not yet demonstrated strong generalization to unseen terrains.

Our work tries to bridge these two. We use a mapping module and a controller module, and train the neural controller end-to-end to generate joint actions from observations. In the controller, our AME-2 encoder also learns planning-level representations (global context and local contact-relevant features) for the subsequent decoder, playing a similar role to the planning module in classical solutions. This design preserves a modular structure at the system level while avoiding explicit online model-based planning. In the mapping module, we keep a simple heuristic structure, but also run a lightweight neural network in the loop to improve robustness and generalization under noise, partial observability, and unseen terrain patterns. 


\subsection{Goal Reaching v.s. Velocity Tracking}

We use a goal-reaching formulation~\cite{nikitagoalreaching} instead of velocity tracking~\cite{he2025attention}. This encourages more agile behaviors and reduces reliance on high-frequency navigation commands. A drawback is that it offers less direct control over the robot’s intermediate motion, which can be troublesome when semantic obstacles lie along the path but appear traversable from the geometry. We envision combining both command interfaces within a single controller in future work, e.g., via masked commands and distillation~\cite{hover}.

\subsection{Whole-Body Skills}

We achieve whole-body locomotion skills in this work, but for a given terrain type the learned motions tend to follow similar contact patterns. For systems with higher DoFs, such as humanoids, the same terrain type at different scales may require different contact strategies. For example, a humanoid may step onto low obstacles with a single leg, use a two-leg jump for medium heights, and leverage both arms and legs for higher obstacles. It remains unclear whether our method, without additional references or priors such as~\cite{yang2025omniretarget}, can automatically discover such diverse contact patterns to handle a broader range of terrains in higher-DoF systems.

{Moreover, whole-body agility and energy efficiency are not always aligned. Some agile motions in our work, such as leaping with active knee contacts on sparse terrain edges, can be energy intensive and unsuitable for applications like industrial inspection. A promising direction is to learn such preferences and command them at deployment, so the policy engages aggressive maneuvers only when necessary.}

\subsection{Towards Higher Success Rates on Unseen Terrains}

We still observe more failures on unseen test terrains than on training terrains, as expected for learning-based systems. Notably, most failures occur during skill transitions. For example, on \textit{Test 2} the robot must first decelerate on a sparse region and then execute a climbing maneuver while maintaining precise hind-leg footholds. While these specific cases could likely be addressed by finetuning~\cite{rudin2025parkour}, it remains unclear whether there is a scalable, principled approach for learning such challenging skill transitions with zero-shot generalization.

\revise{\subsection{Compute Efficiency}
Our pipeline requires 60 RTX-4090-days for ANYmal-D and 30 for TRON1 in Isaac Gym. For comparison, AME-1~\cite{he2025attention} uses 6 A100-days for an ANYmal-D controller that cannot climb high obstacles, and \cite{rudin2025parkour} takes an estimated 30$\sim$40 RTX-4090-days for the full ANYmal-D pipeline without zero-shot generalization. Our pipeline thus trades compute for stronger capabilities and simplicity with less tuning effort. That said, since attention over the entire map dominates the compute cost, sparse attention techniques (e.g., \cite{park2026delta}) could potentially reduce it several fold, yielding a pipeline that is both capable and computationally efficient in the future.
}

\subsection{Limitations and Future Work}

This work has several limitations. First, we use a 2.5-D elevation map and do not address fully 3-D locomotion. Future work could incorporate multi-layer elevation maps~\cite{chen2025learning}, which are directly compatible with our policy architecture, or explore attention-based voxel representations. Second, our controllers are not designed for severely degraded perception, such as high grass or snow. A promising direction is to combine robust controllers~\cite{miki2022learning} with our agile and generalized controllers into a single scene-aware policy. Third, our mapping module can fail in highly dynamic environments with occlusions, and could be extended to explicitly reason about moving elements.

\section*{Acknowledgments}


This work is funded by ETH AI Center, NCCR automation, Apple Inc. Any views, opinions, findings, and conclusions or recommendations expressed in this material are those of the authors and should not be interpreted as reflecting the views, policies or position, either expressed or implied, of Apple Inc.

We thank the help and insights from Yuni Fuchioka, Elena Krasnova, William Talbot, Junzhe He, Matthias Heyrman, Clemens Schwarke, Jiale Fan, Per Frivik, Jonas Frey, Manthan Patel, Emilio Palma, Vladlen Koltun, Matthias Müller, Siyuan Hu. We thank the support from Martin Vechev. We thank the technical support from LimX Dynamics and ANYbotics.

{\appendices

\section{Terrain Parameters}
\label{sec:app_terrain}

We use the following terrain proportions and parameters for training. {AOT: goals are sampled anywhere on the terrain. EOT: goals are sampled at the other end of the terrain.}
\begin{itemize}
    \item \textbf{Rough (Dense, 5\%, {AOT})}: Heightfields generated from uniform noise, with terrain resolution smaller than the map resolution. Over the curriculum, the noise range increases from $\pm0$~m to $\pm0.2$~m for ANYmal-D, and from $\pm0$ m to $\pm0.15$~m for TRON1.
    \item \textbf{Stair Down (Dense, 5\%, {AOT})}: Floating stairs descending with random widths and random walls on both sides. The slope is increased from $5^\circ$ to $45^\circ$ over the curriculum for both robots. 
    \item \textbf{Stair Up (Dense, 5\%, {AOT})}: Floating stairs ascending with random widths and random walls on both sides. The slope is increased from $5^\circ$ to $45^\circ$ over the curriculum for both robots. 
    \item \textbf{Boxes (Dense, 5\%, {AOT})}: Heightfields generated by adding and removing random boxes on flat ground. Over the curriculum, the maximum box height increases from $0.05$~m to $0.4$~m for ANYmal-D, and from $0.05$~m to $0.3$~m for TRON1.
    \item \textbf{Obstacles (Dense, 5\%, {AOT})}: Random positive and negative obstacles placed on random slopes. Over the curriculum, the obstacle density increases from $0~\mathrm{m}^{-2}$ to $0.5~\mathrm{m}^{-2}$ for both robots.
    \item \textbf{Climbing Up (Climbing, 20\%, {EOT})}: A pit to climb out of, with vertices near the edges randomly perturbed by up to 20\% of the height. Over the curriculum, the pit height increases from $0.1$ m to $1.0$ m for ANYmal-D, and from $0.1$ m to $0.48$ m for TRON1. 
    \item \textbf{Climbing Down (Climbing, 5\%, {EOT})}: A platform to climb down from, with vertices near the edges randomly perturbed by up to 20\% of the height. Over the curriculum, the platform height increases from $0.2$ m to $1.0$ m for ANYmal-D, and from $0.2$ m to $0.88$ m for TRON1. 
    \item \textbf{Climbing Consecutive (Climbing, 5\%, {EOT})}: Two stacked layers of rings to climb out of. The robot needs to consecutively climb up twice and then climb down twice. Over the curriculum, the first ring height increases from $0.05$~m to $0.5$~m for ANYmal-D and from $0.05$~m to $0.3$~m for TRON1. The second ring increases from $0.05$~m to $0.4$~m for ANYmal-D and from $0.05$~m to $0.3$~m for TRON1.
    \item \textbf{Gap (Sparse, 5\%, {EOT})}: A gap to jump over, with vertices near the edges randomly perturbed by up to $10\%$ of the distance, and a height difference between the two sides randomly sampled within $\pm30\%$ of the distance. Over the curriculum, the distance increases from $0.1$~m to $1.1$~m for ANYmal-D and to $0.6$~m for TRON1.
    \item \textbf{Pallets (Sparse, 5\%, {EOT})}: Parallel beams with random orientations. Over the curriculum, the beam width decreases from $0.4$~m to $0.16$~m for both robots. The gap width increases from $0.08$~m to $0.35$~m for ANYmal-D and from $0.08$~m to $0.2$~m for TRON1. The inter-beam height difference increases from $0$~m to $0.3$~m for ANYmal-D and from $0$~m to $0.2$~m for TRON1.
    \item \textbf{Stones (Sparse, 30\%, {AOT})}: We follow the ``Stones-Everywhere" design in \cite{zhang2024risky} and \cite{he2025attention}. 
    \item \textbf{Beam (Sparse, 5\%, {EOT})}: A beam connecting two platforms, randomly inclined with roll, pitch, and yaw sampled within $\pm 0.1~\mathrm{rad}$, and with a platform height difference sampled within $\pm0.2$~m. Over the curriculum, the beam width decreases from $0.9$~m to $0.18$~m for both.
    
\end{itemize}

For sparse terrains, we add physical floors to half of them and virtual floors (visible in the map without physical collision) to the other half. The floor height is randomized between $-1.5$~m and $-0.35$~m. This discourages the robot from walking on the floor when traversing shallow sparse terrains.

{
We also present the details of the test terrains below. These test terrains pose non-trivial generalization challenges, as evidenced by the baselines’ substantially lower performance on them than on the training terrains.
}

{
\textbf{Test 1} consists of irregular stepping stones with a more randomized layout. Compared with the training terrain ``Stones", where each stone is placed within a grid and stones are well separated, the stones in Test 1 can overlap and form irregular gaps. It examines generalization on sparse terrains.
}

{
\textbf{Test 2} combines stones, rotated pallets, and high-platform climbing. It requires the robot not only to remain stable on a new combination of sparse terrains, but also to maintain stability while climbing up. 
It examines compositional generalization across combined sparse and climbing terrains.
}

{
\textbf{Test 3} requires the robot to first avoid a large obstacle, which is larger than those seen during training, then climb up, jump over a gap, avoid another large obstacle, and climb down. It examines compositional generalization across obstacle avoidance, agile climbing, and jumping skills.
}

{
\textbf{Test 4} is a realistic debris mesh, where the robot needs to climb over large rubble and avoid getting its feet stuck between pieces. 
It examines generalization to realistic climbing terrains with risky foothold regions.
}

\section{Randomization Parameters}
\label{sec:app_rand}

For both robots, we randomize the payload within $[-5, 5]~\mathrm{kg}$, actuation delays within $[0, 0.02]~\mathrm{s}$, and friction coefficients within $[0.3, 1.0]$. For TRON1, we additionally apply random biases to PD gains and armatures within $\pm 15\%$ of the raw values.

For policy observations, we add zero-mean uniform noise with the following maximum magnitudes: $0.1$~m/s for base linear velocity, $0.2$~rad/s for base angular velocity, $0.05$ for projected gravity, $0.01$~rad for joint positions, $1.5$~rad/s for joint velocities, and $0.05$~m for map observations. We also add a random drift within $[-0.03, 0.03]~\mathrm{m}$ when querying map observations.

During student policy training, we make $15\%$ of depth cloud points missing and $2\%$ artifacts. We give $10\%$ of the environments access to complete maps with variance $0.0025~ \mathrm{m}^2$. We corrupt $1\%$ points in the student map observations by assigning random values with a random variance larger than $1~\mathrm{m}^2$.

\section{PPO Parameters}
\label{sec:app_ppo}

We use the following PPO parameters in \Table{ppoparam}.

\begin{table}[ht]
    \centering
    \caption{PPO Parameters}
    \label{tab:ppoparam}
    \resizebox{.95\columnwidth}{!}{%
    \begin{tabular}{ll} \hline
        Parameter {(RSL-RL~\cite{schwarke2025rsl} Conventions)} & Value \\ \hline
        Value Loss Coefficient &  1.0 \\
        Value Loss Clip &  0.2 \\
        Entropy Coefficient & Decay from 0.004 to 0.001 \\
        Learning Epochs per Iteration & 4 \\
        Mini Batches per Iteration & 3 \\
        Simulation Steps per Iteration & 24 \\
        Number of Environments & 4800 \\
        Learning Rate & Adaptive  \\
        $\gamma$ & 0.99 \\
        $\lambda$ & 0.95 \\
        Desired KL Divergence & 0.01 \\ \hline
        \textit{Student only} & \\
        Learning Rate When Surrogate Loss Disabled & 0.001 \\
        Action Distillation Loss Coefficient & 0.02 \\
        Representation Loss Coefficient & 0.2 \\\hline
    \end{tabular}}
    \vspace{-3mm}
\end{table}

\section{Temporal Recurrent Baseline Mapping Model }
\label{sec:t_rec_map}

The architecture of the recurrent model used as the mapping baseline is shown in \fig{tmprec_baseline}. 
We provide the model with ground-truth delta transforms, which are sufficient for map reconstruction and serve as a clean replacement for the proprioceptive observations used in prior work~\cite{sun2025dpl, yu2024walking}, where transforms are not directly available. {This model has 6M parameters and takes 1.76 ms per step for 1000 parallel ANYmal environments during training.} 

\begin{figure}[ht]
    \centering
    \includegraphics[width=0.9\linewidth]{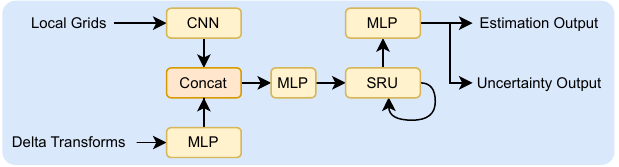}
    \vspace{-2.5mm}
    \caption{The temporal recurrent model architecture to predict egocentric elevation mapping. {The CNN has two 4-channel layers with 5×5 kernels, followed by a linear mapping ($1\times1$ convolution after flattening) to a 512-d embedding. The delta-transform embedding MLP has two 32-unit layers, the intermediate MLP has one 512-unit layer, the decoder MLP has one 1024-unit layer, and the SRU is a one-layer LSTM variant with hidden dimension 512.}}
    \label{fig:tmprec_baseline}
    \vspace{-0.5cm}
\end{figure}
}

\bibliographystyle{IEEEtran}
\bibliography{IEEEabrv,references}
%


 




\vfill

\end{document}